\documentclass[sigconf]{acmart}

\usepackage{microtype}
\usepackage{tabularx}
\usepackage{booktabs}
\usepackage{multirow}
\usepackage{color}
\usepackage{bm}
\usepackage{graphicx}
\usepackage{algorithm}
\usepackage{algorithmicx}
\usepackage{balance}
\usepackage{CJK}
\usepackage{url}
\usepackage{algpseudocode}
\algnewcommand\algorithmicinput{\textbf{Input:}}
\algnewcommand\INPUT{\item[\algorithmicinput]}
\algnewcommand\algorithmicoutput{\textbf{Output:}}
\algnewcommand\OUTPUT{\item[\algorithmicoutput]}

\AtBeginDocument{%
  \providecommand\BibTeX{{%
    \normalfont B\kern-0.5em{\scshape i\kern-0.25em b}\kern-0.8em\TeX}}}

\copyrightyear{2021}
\acmYear{2021}
\setcopyright{acmcopyright}
\acmConference[CIKM '21]{Proceedings of the 30th ACM International Conference on Information and Knowledge Management}{November 1--5, 2021}{Virtual Event, QLD, Australia}
\acmBooktitle{Proceedings of the 30th ACM International Conference on Information and Knowledge Management (CIKM '21), November 1--5, 2021, Virtual Event, QLD, Australia}
\acmPrice{15.00}
\acmDOI{10.1145/3459637.3482352}
\acmISBN{978-1-4503-8446-9/21/11}

\begin{document}
\fancyhead{}

\title{Identifying Untrustworthy Samples: Data Filtering for Open-domain Dialogues with Bayesian Optimization}



\author{Lei Shen$^{1,2}$,\quad Haolan Zhan$^2$,\quad Xin Shen$^3$,\quad Hongshen Chen$^{4}$,\quad Xiaofang Zhao$^{1}$*,  \quad Xiaodan Zhu$^5$}


\makeatletter
\def\authornotetext#1{
\if@ACM@anonymous\else
    \g@addto@macro\@authornotes{
    \stepcounter{footnote}\footnotetext{#1}}
\fi}
\makeatother
\authornotetext{Corresponding author.}


\affiliation{
 \institution{\textsuperscript{\rm 1}Institute of Computing Technology, Chinese Academy of Sciences, Beijing, China}
 \institution{\textsuperscript{\rm 2}University of Chinese Academy of Sciences, Beijing, China \qquad \textsuperscript{\rm 3}Australian National University \country{Australia}}
 \institution{\textsuperscript{\rm 4}JD.com, Beijing, China \qquad \textsuperscript{\rm 5}Queen's University \country{Canada}}
 }
\email{shenlei17z@ict.ac.cn, zhanhaolan316@gmail.com, u6498962@anu.edu.au, ac@chenhongshen.com, zhaoxf@ict.ac.cn}

\def\authors{Lei Shen, Haolan Zhan, Xin Shen, Hongshen Chen, Xiaofang Zhao, Xiaodan Zhu}

\renewcommand{\shortauthors}{Shen et al.}



\begin{abstract}
Being able to reply with a related, fluent, and informative response is an indispensable requirement for building high-quality conversational agents. In order to generate better responses, some approaches have been proposed, such as feeding extra information by collecting large-scale datasets with human annotations, designing neural conversational models (NCMs) with complex architecture and loss functions, or filtering out \textit{untrustworthy} samples based on a dialogue attribute, e.g., Relatedness or Genericness. In this paper, we follow the third research branch and present a data filtering method for open-domain dialogues, which identifies untrustworthy samples from training data with a quality measure that linearly combines seven dialogue attributes. The attribute weights are obtained via Bayesian Optimization (BayesOpt) that aims to optimize an objective function for dialogue generation iteratively on the validation set. Then we score training samples with the quality measure, sort them in descending order, and filter out those at the bottom. Furthermore, to accelerate the ``filter-train-evaluate'' iterations involved in BayesOpt on large-scale datasets, we propose a training framework that integrates maximum likelihood estimation (MLE) and negative training method (NEG). The training method updates parameters of a trained NCMs on two small sets with newly maintained and removed samples, respectively. Specifically, MLE is applied to maximize the log-likelihood of newly maintained samples, while NEG is used to minimize the log-likelihood of newly removed ones. Experimental results on two datasets show that our method can effectively identify untrustworthy samples, and NCMs trained on the filtered datasets achieve better performance.

\end{abstract}







\maketitle

\section{Introduction}
\label{sec:intro}

With the availability of large-scale data, such as posts on social media or forums, scripts of movies or TV series, and datasets from collection or crowdsourcing \cite{lison2016opensubtitles2016,henderson2019repository,chen2020jddc}, neural conversational models (NCMs) \cite{shen2019modeling,shen2021icassp} have developed rapidly and their performance has been improved steadily. Current NCMs are mainly trained on context-response pairs\footnote{A pair of context and response is a training sample in the dialogue generation task.} with maximum likelihood estimation (MLE). However, the generated responses may suffer from some notorious problems, such as being generic, inconsistent, or unrelated to the given contexts. Previous studies tried to solve these issues by feeding extra information, e.g., topics \cite{liu2020nlpcc}, sentence types \cite{reed2018can}, personas \cite{li2016persona}, emotions \cite{zhou2018emotional,shen2020cdl}, documents \cite{meng2019refnet}, multi-modal \cite{shen2021acmmm} or knowledge \cite{ke2018generating,zhan2021naacl,zhou2018commonsense,niu2019knowledge,zhan2021augmenting}, augmenting the model itself \cite{zhao2017learning,xu2018better}, or modifying the loss function \cite{li2016diversity}. 

\begin{table}[!t]
\centering
\caption{Examples of untrustworthy samples from two open-domain dialogue datasets: OpenSubtitles (Case 1 and 2) and DailyDialog (Case 3). ``$c$" and ``$r$" represent context and response, respectively. ``$\rightarrow$" denotes the concatenation of consecutive utterances.}
\vspace{-1mm}
\begin{tabular}{l|lp{6.5cm}}
  \toprule[1pt]
  \multirow{2}*{1} & $c$ & A big nail should be put in your head. \\ \cline{2-3}
  & $r$ & Who are they? (\textcolor[rgb]{0.5,0,0}{An unrelated response}) \\
  \midrule[1pt]
  \multirow{3}*{2} & $c$ & We met yesterday. $\rightarrow$ Oh, you're Thomas. $\rightarrow$ Yes. \\ \cline{2-3}
  & $r$ & We haven't met. (\textcolor[rgb]{0.5,0,0}{An inconsistent response})\\
  \midrule[1pt]
  \multirow{4}*{3} & $c_1$ & There! You can see a window there. \\ \cline{2-3}
  & $c_2$ & It's about an hour. \\ \cline{2-3}
  & $c_3$ & You can buy a bus schedule in a news stand. \\ \cline{2-3}
  & $r$ & I see. Thank you. (\textcolor[rgb]{0.5,0,0}{A generic response})\\ 
  \bottomrule[1pt]
\end{tabular}
\label{tab:case_for_intro}
\end{table}

In addition to the above mentioned approaches, some researchers pay attention to data filtering methods that remove \textit{untrustworthy} training samples and further improve the performance of open-domain dialogue generation. Take Table \ref{tab:case_for_intro} as an example. The response in Case 1 is unrelated to the given context, while the response in Case 2 is inconsistent to the previous utterances. In Case 3, the response can be used to respond three different contexts, and is considered as a generic response. When it comes to the question ``how to identify \textit{untrustworthy} samples?'', each work firstly defines a measure based on an individual dialogue attribute, including Coherence \cite{xu2018better}, Genericness \cite{csaky2019improving}, Relatedness and Connectivity \cite{akama2020filtering}\footnote{The definition of \textit{Relatedness} from \citet{akama2020filtering} is the same as that of \textit{Coherence} from \citet{xu2018better}. Though \citet{akama2020filtering} used both \textit{Relatedness} and \textit{Connectivity}, these two attributes are combined simply with a weighted sum, where the weights are calculated by overall scores on the training set.}. Then samples with low (or high) scores\footnote{For metrics like \textit{Coherence}, lower is worse; while for metrics like \textit{Genericness}, higher is worse.} are regarded as \textit{untrustworthy} ones. However, due to the subjectivity and open-ended nature of human conversations, the quality of dialogue data varies greatly \cite{cai2020learning}. Moreover, it is hard to evaluate the dialogue quality just grounding on a single metric \cite{mehri2020unsupervised,mehri2020usr,pang2020towards}. In general, a reasonable assessment should contain multiple aspects of attributes \cite{see2019makes}, such as the fluency and specificity of the response, the topical relatedness and factual consistency between the context and response, etc.

In this paper, we propose a data filtering method to identify untrustworthy samples in training data with the consideration of seven dialogue attributes. Instead of using these attributes in isolation, we gather them together to learn a measure of data quality based on Bayesian Optimization (BayesOpt) \cite{brochu2010tutorial}. Specifically, the measure is a combination of seven attributes and the weights are learnt via BayesOpt. BayesOpt aims to iteratively find a group of weights that optimizes an objective function on the validation set for dialogue generation. After obtaining the optimal weights and calculating the measure score for each sample, we can filter out data with low scores and use the maintained (filtered) dataset\footnote{In this paper, ``maintained dataset'', ``filtered dataset'', and $M_t$ have the same meaning.} to train an NCM. Nevertheless, simply applying BayesOpt for data filtering on large-scale dialogue datasets is inappropriate, since an iteration of ``filter-train-evaluate'' based on lots of samples is time-consuming. To accelerate the process, we design a training framework, named Diff-MLE-NEG, which combines maximum likelihood estimation (MLE) and negative training method (NEG) \cite{he2019negative}. Given a model trained on a randomly filtered dataset, Bayesian Optimization with Diff-MLE-NEG is performed to update the model parameters iteratively. In each iteration, we keep track of a small size of newly maintained or removed sample set. For newly maintained samples, denoted as desirable ones, MLE is applied to maximize their log-likelihood, while for newly removed samples, denoted as undesirable ones, NEG is used to minimize their log-likelihood.

Experimental results on two open-domain dialogue datasets, OpenSubtitles and DailyDialog, show that our data filtering method can effectively detect untrustworthy samples. Furthermore, after training several NCMs on our filtered datasets, the performance of dialogue generation can be improved.

Our main contributions can be summarized as: (1) We propose a measure based on seven dialogue attributes for estimating the quality of dialogue samples in an unsupervised manner. (2) We frame the dialogue data filtering task as an optimization problem, and propose a new method to accelerate the iterations of ``filter-train-evaluate'' involved in Bayesian Optimization on large-scale datasets. (3) Automatic and human evaluations show that our method can identify untrustworthy dialogue samples and further improve the performance of several NCMs for dialogue generation on different datasets.

\section{Data filtering Method}
\label{sec:method}
In this section, we introduce our proposed data filtering method for open-domain dialogues in detail. We firstly give a description of the task, and then illustrate the Bayesian Optimization approach used for data filtering as well as the definitions of two important components, i.e., dialogue attributes and the objective function. Finally, we design a training method that combines maximum likelihood optimization (MLE) and negative training method (NEG) to accelerate the optimization iterations performed on large-scale datasets.

\subsection{Task Definition}
Given a dialogue corpus $X$ with $N$ pairs of context $c$ and response $r$, i.e., $X = \{(c_i, r_i)\}_{i=1}^N$, data filtering aims to remove untrustworthy utterance pairs from the training data, and further improve the performance of NCMs for dialogue generation \cite{akama2020filtering}. 

To score each sample ($c$-$r$ pair), we propose a measure $S$ as a linear combination of several dialogue attributes:
\begin{equation}
    S = w^{\intercal} \varphi(X),
\label{eq:S}
\end{equation}
where $\varphi(X) \in \mathbb{R}^{n \times 1}$ is the dialogue attributes further described in Section \ref{sec:attributes} for each training sample, $n$ is the number of attributes, and $w \in \mathbb{R}^{n \times 1}$ denotes the weights we want to obtain for these attributes. After each sample is scored according to the measure $S$, we consider pairs with low scores as untrustworthy samples and filter them out for the dialogue generation task. We define the objective function $J$ as the automatic evaluation metric for dialogue generation performed on the validation set. Then we convert the dialogue data filtering task into an optimization problem, and our goal is to define $\varphi(X)$ and find the weights $w$ that optimizes $J$.

\subsection{Bayesian Optimization for Data Filtering}
The defined measure $S$ is agnostic of the objective function $J$, thus we cannot apply gradient-based methods for optimization. Other search methods such as random search or grid search requires exponential traverses regarding the number of parameterization of $w$, which is time-consuming and not suitable for our task.

Inspired by \citet{tsvetkov2016learning} and \citet{ruder2017learning}, we use Bayesian Optimization \cite{brochu2010tutorial} here, which is a framework used to globally optimize any black-box function \cite{shahriari2015taking}. Bayesian Optimization can be considered as a sequential approach to performing a regression from high-level model parameters (e.g., hidden state dimension, the number of layers in a neural network, or $w$ in our method) to the loss function or the objective function \cite{tsvetkov2016learning}.

\begin{figure}[t]
\begin{center}
   \includegraphics[width=0.95\linewidth]{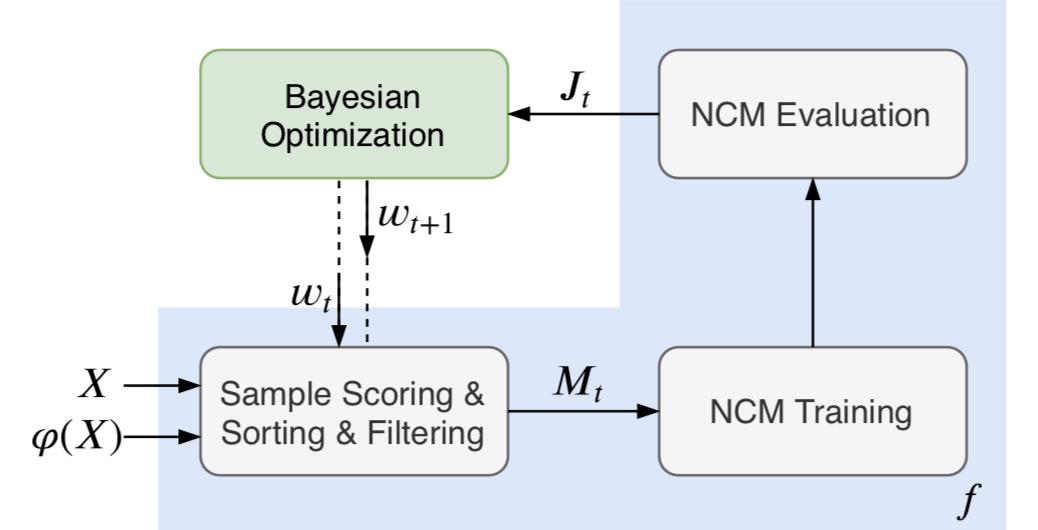}
\end{center}
\vspace{-2mm}
   \caption{Overview of data filtering with Bayesian Optimization. ``$X$" and ``$\varphi(X)$" represent the training set and dialogue attributes, respectively. ``$M$'' denotes the maintained/filtered dataset. The black-box function $f$ takes weights $w$ as input and returns the objective function $J$, while Bayesian Optimization finds the optimal $w$ that minimizes $J$ iteratively.}
\label{fig:overview}
\end{figure}

In general, given a black-box function $f: \mathbb{X} \rightarrow \mathbb{R}$, Bayesian Optimization tries to find an input $x' \in \mathop{\arg\min}\nolimits_{x \in X} f(x)$ that globally minimize $f$. To fulfill this, it requires a prior $p(f)$ over the function and an acquisition function $a_{p(f)}: \mathbb{X} \rightarrow \mathbb{R}$ that calculates the utility of any evaluation at any $x$. Bayesian Optimization then proceeds in an iterative manner. At iteration step $t$, (1) it obtains the most promising input $x_t \in \mathop{\arg\max}a_{p(f)}(x)$ via numerical optimization; (2) then, it evaluates the surrogate function $y_t \sim f(x_t) + \mathcal{N}(0, \sigma^2)$ on input $x_t$, and adds the resulting data point ($x_t$, $y_t$) to the set of observations $\mathcal{O}_{t-1}=\{(x_j, y_j)\}_{j=1}^{t-1}$; (3) based on $\mathcal{O}_t$, it updates the prior $p(f|\mathcal{O}_t)$ and the acquisition function $a_{p(f|\mathcal{O}_t)}$. Following \citet{ruder2017learning}, Gaussian Processes (GP) is chosen for $p(f)$, and GP with Monte Carlo acquisition and Expected Improvement (EI) \cite{mockus1975bayes} is used as the acquisition function. 
Note that the Robust Bayesian Optimization framework, RoBO\footnote{\url{http://automl.github.io/RoBO/}}, implements several details in Bayesian Optimization, and can be applied directly.

As for utilizing Bayesian Optimization in our task, the data filtering process described in Figure \ref{fig:overview} can be treated as the black-box function: (1) it takes $w$ that should be evaluated as input; (2) all training samples are sorted according to the measure $S$ defined in Equation \ref{eq:S}; (3) we remove the lowest $n\%$ samples and get a subset of $X$, denoted as $M$; (4) a specific NCM is trained on $M$; (5) the model is evaluated on the validation set according to the evaluation metric $J$ and the value of $J$ is returned. 

\subsection{$\varphi(X)$ and $J$}
\label{sec:attributes}
$\varphi(X)$ in Equation \ref{eq:S} denotes the values of some dialogue attributes. There exists several work focusing on dialogue attributes, and some use these features to influence the training and inference processes of dialogue models \cite{see2019makes,zhang2018generating,cai2020learning}, while others apply them as metrics for evaluation \cite{tao2018ruber,zhao2020designing,pang2020towards,mehri2020unsupervised,mehri2020usr}. Our paper follows the former branch, and we use seven dialogue attributes to calculate the measure $S$.

\noindent\textbf{Specificity.}
Generic and unspecific responses are common in open-domain datasets, e.g., ``I don't know'' or ``That's great''. Following \citet{see2019makes}, we apply Normalized Inverse Document Frequency (NIDF) to measure word rareness:
\begin{equation}
    \mathrm{NIDF}(w) = \frac{\mathrm{IDF}(w)-\mathrm{idf}_{min}}{\mathrm{idf}_{max} - \mathrm{idf}_{min}},
\end{equation}
where $\mathrm{IDF}(w) = \mathrm{log}\frac{N_r}{N_{rw}}$. $N_r$ is the number of responses in the dataset, and $N_{rw}$ is the number of responses that contain word $w$. $\mathrm{idf}_{min}$ and $\mathrm{idf}_{max}$ are the minimum and maximum $\mathrm{IDFs}$, taken over all words in the vocabulary. Then, the specificity of a response $r$, $\mathrm{Spec}(r)$, is the mean $\mathrm{NIDF}$ of all words in $r$.

\noindent\textbf{Repetitiveness.}
Specificity cares more about inter-utterance difference, for intra-utterance diversity, we use $\mathrm{Rept}(r)$ \cite{cai2020learning} to measure the repetitiveness of a response $r$: 
\begin{equation}
    \mathrm{Rept}(r) = \frac{\sum_{i=1}^{|r|} \mathrm{I}(w_i \in \{w_0, ..., w_{i-1}\})}{|r|},
\end{equation}
where $\mathrm{I}(\cdot)$ is an indicator function that takes the value 1 when $w_i \in \{w_0, ..., w_{i-1}\}$ is true and 0 otherwise. $|r|$ is the length of $r$.

\noindent\textbf{Relatedness.}
Cosine similarity between the vectors of two utterances, $c$\footnote{For contexts with multiple utterances, we concatenate them into a long string.} and $r$, has been widely used to calculate their topical relatedness \cite{zhang2018generating, xu2018better,akama2020filtering}:
\begin{equation}
    \mathrm{Rel}(c, r) = \mathrm{cos}(v(c), v(r)), 
\end{equation}
where $v(\cdot)$ is the utterance embedding, and is computed as the weighted sum of word embeddings: $v(s) = \frac{1}{|s|}\sum_{w \in s} \frac{0.001}{0.001+p(w)} e(w)$ \cite{arora2016simple}, where $e(w)$ and $p(w)$ are the embedding and probability\footnote{Probability is calculated based on the maximum likelihood estimation on the training data.} of word $w$, respectively.

\begin{figure*}[t]
\begin{center}
   \includegraphics[width=0.95\linewidth]{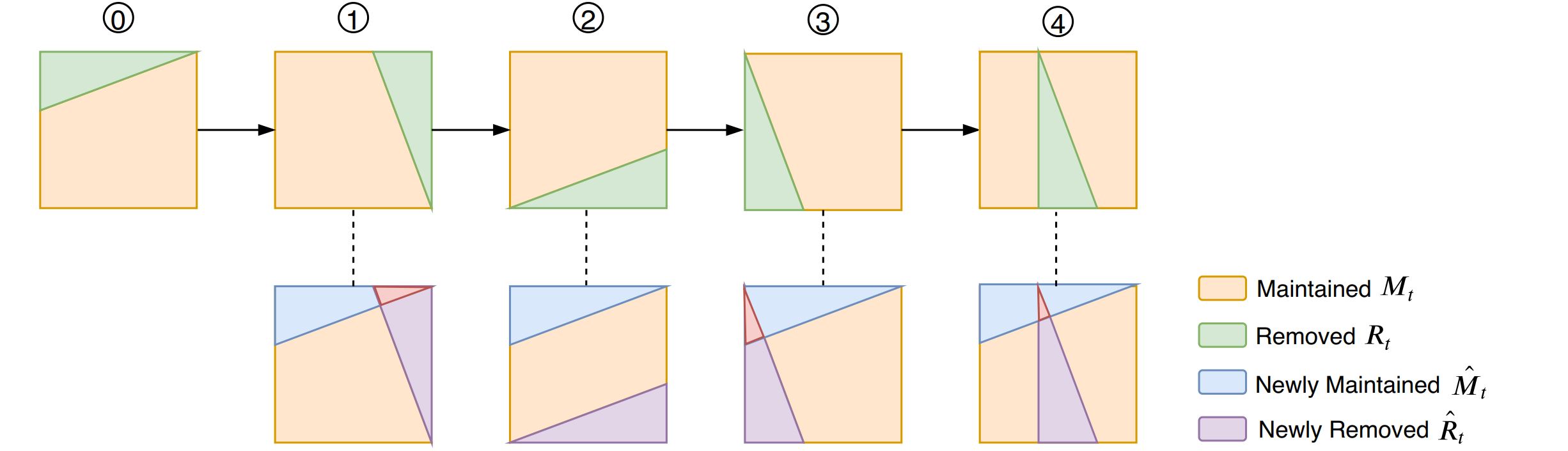}
\end{center}
\vspace{-2mm}
   \caption{Illustration of the training acceleration process. $M_0$ and $R_0$ are obtained by randomly removing bottom $n\%$ samples in $X$. An NCM is trained on $M_0$ to convergence, and evaluated on the validation set to get $J_0$. The maintained sample sets $\{M_t\}_{t=1}^k$ and removed sample sets $\{R_t\}_{t=1}^k$ are the outputs of ``score-sort-filter'' process, where the weights $w_t$ in measure score $S$ are computed by Bayesian Optimization based on $J_{t-1}$. The newly maintained sample sets $\{\hat{M}_t\}_{t=1}^k$ are the difference sets of $\{M_t\}_{t=1}^k$ and $M_0$, respectively. The newly removed sample sets $\{\hat{R}_t\}_{t=1}^k$ are the difference sets of $\{R_t\}_{t=1}^k$ and $R_0$, respectively. $k$ is the iteration number of Bayesian Optimization.} 
\label{fig:accelerate}
\end{figure*}

\noindent\textbf{Continuity.}
A good open-domain conversational agent should be able to interact with users in more turns, that is, a response is also responsible to encourage the next utterance. Following \citet{cai2020learning}, $\mathrm{Cont}(r, u)$ is introduced as the cosine similarity between vectors of $r$ and the subsequent utterance $u$ with the same calculation as Relatedness. 

\noindent\textbf{Coherence.}
Coherence evaluates whether a response can be considered as an appropriate and natural reply to the context, and it is also described as Connectivity \cite{akama2020filtering}. Coherence between $c$ and $r$ can be either measured as next sentence prediction \cite{tao2018ruber} or conditional language modeling task \cite{pang2020towards}. With the support of large-scale pre-trained language models, e.g., GPT-2 \cite{radford2019language}, the latter one could be more reliable to capture the coherence between $c$ and $r$ after we further fine-tune GPT-2 on dialogue datasets. Following \citet{pang2020towards}, Coherence is defined as follows (ranging from 0 to 1):
\begin{equation}
    \mathrm{Coh}(c, r) = - \frac{\mathrm{max}(\mathrm{C}_5, P_G(r|c))-\mathrm{C}_5}{\mathrm{C}_5}, 
\end{equation}
where $P_G$ represents the fine-tuned GPT-2 model, and $P_G(r|c) = \frac{1}{|r|}\sum_{t=1}^{|r|} \mathrm{log} P_G(r_t|r_{<t}, c)$. Since $P_G(r|c)$ is a negative unbounded number, a lower bound $\mathrm{C_5}$ used to normalize Coherence is defined as 5th percentile of the score distribution. Please refer to \citet{pang2020towards} for more details.

\noindent\textbf{Fluency.}
Fluency assesses the grammatical correctness and readability of an utterance. Similar to Coherence, the fluency score is computed as:
\begin{equation}
    \mathrm{Flu}(r) = - \frac{\mathrm{max}(\mathrm{F}_5, P_G(r))-\mathrm{F}_5}{\mathrm{F}_5},
\end{equation}
where $P_G(r) = \frac{1}{|r|}\sum_{t=1}^{|r|}\mathrm{log}P_G(r_t|r_{<t})$, and $\mathrm{F_5}$ denotes 5th percentile of the score distribution.

\noindent\textbf{Consistency.}
Factual consistency is also an important attribute for a dialogue, and there should not be logic contradictions. Natural Language Inference (NLI) \cite{bowman2015large} tries to map a sentence pair to an entailment category including ''Entailment'', ``Neutral'', and ``Contradiction'' (abbreviated as ``Contra'' below). Besides, it has been used in dialogue systems to measure the logic consistency \cite{welleck2019dialogue,song2020generating,pang2020towards}. We apply the pre-trained model RoBERTa\footnote{We use RoBERTa released by Huggingface: \url{https://huggingface.co/roberta-large-mnli}.} \cite{liu2019roberta}, $P_R$, to compute the consistency score between $c$ and $r$ as follows:
\begin{equation}
    \mathrm{Cons}(c, r) = 1-P_R(``\mathrm{Contra}"|c, r).
\end{equation}

\noindent\textbf{Choices of $J$.} For the objective function $J$, we mainly try two simple options: perplexity \cite{serban2015hierarchical} and the negative value\footnote{Bayesian Optimization needs to minimize $J$, but most metrics we used for evaluating dialogue generation should be larger when indicating a better response.} of a sum of 13 widely-used automatic metrics, including BLEU, Dist-1/2/3, Intra-1/2/3, Ent-1/2, Average, Greedy, Extrema, and Coherence \cite{xu2018better}. For the latter one, please refer to \citet{cai2020learning} for more details. For simplicity, we denote these two types of $J$ as ``+ppl'' and ``-metric'', respectively. Some automatic metrics \cite{tao2018ruber,zhao2020designing,sinha2020learning,pang2020towards,mehri2020unsupervised,mehri2020usr} that correlate more with human judgment have been proposed recently, but they are not accepted thoroughly by the entire community. Therefore, we leave applying these metrics as $J$ for future work.

\subsection{Training Acceleration}

As we can see in Figure \ref{fig:overview}, in each iteration of Bayesian Optimization, we need to repeat filtering, training and evaluation. Since the maintained dataset $M_t$ is still large, training an NCM with it is extremely time-consuming. To accelerate the process, we design a training method, Diff-MLE-NEG, which combines maximum likelihood estimation (MLE) and negative training method (NEG) \cite{he2019negative} together. 

Following \citet{he2019negative}, MLE is used to maximize the log-likelihood of desirable samples, while NEG aims to minimize the log-likelihood of undesirable ones.
Here the desirable samples are defined as newly maintained ones in iteration $t$, i.e., $\hat{M}_t=M_t-M_0$, while the undesirable samples are defined as newly removed ones in iteration $t$, i.e., $\hat{R}_t=R_t-R_0$, which is illustrated in Figure \ref{fig:accelerate}\footnote{We use the difference set of $M_t$ (or $R_t$) and $M_0$ (or $R_0$), rather than $M_t$ (or $R_t$) and $M_{t-1}$ (or $R_{t-1}$) (the difference set of two consecutive iterations), to achieve a stable update process.}. $M_0$ and $R_0$ are obtained by randomly removing bottom $n\%$ samples in $X$. An NCM is trained to convergence with $M_0$ at first, and then Diff-MLE-NEG (shown in Algorithm \ref{alg:neg}) updates the model parameter $\theta$ on $\{\hat{M}_t\}_{t=1}^k$ and $\{\hat{R}_t\}_{t=1}^k$ iteratively, where $k$ is the number of optimization iterations. 

\begin{algorithm}[t] 
\caption{Diff-MLE-NEG} 
\label{alg:neg} 
\begin{algorithmic}[1]
\INPUT NCM parameter $\theta$, newly maintained samples $\hat{M}_t$, and newly removed samples $\hat{R}_t$
\State \Comment{MLE update}
\For{$\langle x_M, y_M \rangle$ in $\hat{M}_t$}
\State $\theta = \theta + \alpha \cdot \nabla_\theta \mathrm{log}P_\theta(y_M|x_M)$
\EndFor
\State \Comment{NEG update}
\For{$\langle x_R, y_R \rangle$ in $\hat{R}_t$}
\State $\theta = \theta - \alpha \cdot \nabla_\theta \mathrm{log}P_\theta(y_R|x_R)$
\EndFor
\end{algorithmic}
\end{algorithm}

\section{Experiments}
\label{sec:experiments_settings}
In this section, we introduce some empirical settings of our experiments, including research questions, datasets, models and comparable methods, implementation details, and evaluation measures.

\subsection{Research Questions}
We aim at answering the following research questions:

\noindent (\textbf{RQ1}): What is the performance of our approach on automatic and human evaluations? Does our model outperform other comparable methods?

\noindent (\textbf{RQ2}): Is our method able to identify untrustworthy samples, and even to sort training samples properly?

\noindent (\textbf{RQ3}): Does our data filtering method improve the generation quality and generate better response?

\noindent (\textbf{RQ4}): What are the relations among chosen attributes? Are they reasonable to be included in measure $S$?

\noindent (\textbf{RQ5}): Where does the improvements of our method come from? What is the difference between using only one attribute and our proposed measure $S$?

\noindent (\textbf{RQ6}): How does Bayesian Optimization work in this method?

\begin{table}[!t]
    \centering
    \caption{Statistics of two experimental datasets, OpenSubtitles and DailyDialog. ``Avg.'' is the abbreviation of average, and ``$\#$'' denotes ``number''.}
    \vspace{-2mm}
    \begin{tabular}{l|c|c} 
    \toprule[1pt]
        & OpenSubtitles & DailyDialog \\ \hline
        Training size & 298,780 & 54,889 \\ 
        Validation size & 29,948 & 6,005 \\ 
        Test size & 29,940 & 5,700 \\ 
        Vocabulary size & 116,611 & 17,875 \\ 
        Avg. $\#$ of word per utterance & 12.89 & 14.87 \\
    \bottomrule[1pt]
    \end{tabular}
    \label{tab:training_set_statistics}
\end{table}

\subsection{Datasets}
Following \citet{csaky2019improving} and \citet{akama2020filtering}, we conduct experiments on two widely-used open-domain dialogue datasets: OpenSubtitles \cite{lison2016opensubtitles2016} and DailyDialog \cite{li2017dailydialog}. OpenSubtitles is a collection of movie subtitles and originally contains over 2 billion utterances. DailyDialog consists of 90,000 utterances in 13,000 daily conversations. After data preprocessing\footnote{We use the preprocessed datasets provided by \citet{cai2020learning}.}, including tokenization and utterance concatenation, the number of $c$-$r$ pairs in training/validation/test set is 298,780/29,948/29,940 for OpenSubtitles, and 54,889/6,005/5,700 for DailyDialog. The detailed statistics of two datasets we used in experiments are shown in Table \ref{tab:training_set_statistics}. As we can see in Table \ref{tab:case_for_intro}, both of these datasets include some untrustworthy $c$-$r$ pairs.

\subsection{Models and Comparable Methods}
We perform experiments using four representative NCMs for open-domain dialogue generation:
\begin{itemize}
    \item S2S \cite{bahdanau2014neural}: a sequence-to-sequence model with cross-attention mechanism.
    \item CVAE \cite{zhao2017learning}: a conditional variational auto-encoder model with KL-annealing and a bag-of-word (BOW) loss.
    \item TRS \cite{vaswani2017attention}: the Transformer model, which is an encoder-decoder architecture with self-attention mechanism.
    \item GVT \cite{lin2020variational}: a Transformer model with a global latent variable.
\end{itemize} 
We also compare our method with an entropy-based filtering approach \cite{csaky2019improving}, which aims to remove generic utterances from the training data for promoting less-safe response generation. In our experiments, we apply IDENTITY-BOTH\footnote{IDENTITY means the entropy is calculated on all utterances, while BOTH means the filtering is conducted on both context and response sides.} method of \citet{csaky2019improving} and denote it as ``-Ent'' to show the results.

\subsection{Implementation Details}
We employ a 2-layer bidirectional GRU \cite{cholearning} as the encoder and a unidirectional one as the decoder for S2S and CVAE. The hidden size is set to 300, and the size of latent variable is set to 64 for CVAE. For TRS and GVT, the hidden size, number of attention heads and number of hidden layers are set to 300, 2, and 2, respectively. The size of latent variable for GVT equals to 300. The word embedding is initialized with the 300-dimensional pre-trained GloVe embeddings \cite{pennington2014glove} for both encoder and decoder. KL annealing and the BOW loss are applied as in \citet{zhao2017learning}. $k$, the number of Bayesian Optimization iteration, is set to 100. The proportion ($n$\%) of training samples we need to filter out is 26\% and 12\% for OpenSubtitles and DailyDialog, respectively\footnote{\textbf{In order to make fair comparisons, the proportion number is determined by IDENTITY-BOTH method in the work of \citet{csaky2019improving}.}}. In the test time, we use greedy decoding strategy for all models. Each model is trained on two kinds of datasets: the original dataset and the filtered dataset obtained from our data filtering method, keeping other configurations the same. To restrict the NEG training method and avoid the loss being too large, we utilize gradient penalty following \citet{gulrajani2017improved}. All the models are implemented with Pytorch\footnote{\url{https://pytorch.org/}} and are trained on four Titan Xp GPUs.

\begin{table*}[!t]
    \centering
    \caption{Automatic evaluation results (\%) on the test set of OpenSubtitles. ``-Orig'' means models are trained on the original (non-filtered) dataset, while ``-Ent'' represents the IDENTITY-BOTH method of \citet{csaky2019improving}. ``$\spadesuit$'' and ``$\heartsuit$'' denote that the model is trained on filtered datasets with $J$ = ``+ppl" and $J$ = ``-metric'', respectively. Bold face indicates the best result in terms of the corresponding metric.}
    \vspace{-2mm}
    \resizebox{\textwidth}{34mm}{
    \begin{tabular}{lcccccccccccc}
    \toprule[1pt]
        Model & BLEU & Perplexity$\downarrow$ & Dist-1 & Dist-2 & Dist-3 & Intra-1 & Intra-2 & Intra-3 & Average & Greedy & Extrema & Coherence \\
    \midrule[1pt]
        S2S-Orig & \textbf{0.55} & \textbf{46.59} & 0.13 & 0.54 & 1.31 & 98.65 & 99.42 & 99.59 & 84.95 & 73.63 & 54.80 & 89.29 \\
        S2S-Ent & 0.41 & 57.79 & 0.07 & 0.27 & 0.52 & 97.76 & 98.23 & 98.62 & \textbf{87.87} & \textbf{75.59} & 54.45 & \textbf{92.41} \\
        \textbf{S2S (}$\spadesuit$\textbf{)} & 0.52 & 47.96 & 0.14 & 0.59 & 1.26 & 99.05 & 99.47 & 99.68 & 85.04 & 73.69 & 55.08 & 88.25 \\ 
        \textbf{S2S (}$\heartsuit$\textbf{)} & 0.54 & 47.88 & \textbf{0.17} & \textbf{0.62} & \textbf{1.36} & \textbf{99.12} & \textbf{99.52} & \textbf{99.73} & 85.13 & 73.71 & \textbf{55.14} & 90.75 \\ \hline
        CVAE-Orig & 0.21 & \textbf{37.78} & 0.07 & 0.28 & 0.62 & 99.28 & 99.46 & 99.79 & 86.97 & 73.72 & 53.41 & 91.02 \\ 
        CVAE-Ent & 0.19 & 41.51 & 0.06 & 0.31 & 0.70 & 97.90 & 98.60 & 98.80 & 87.12 & 72.30 & 51.60 & \textbf{91.27} \\
        \textbf{CVAE (}$\spadesuit$\textbf{)} & 0.20 & 40.36 & 0.09 & 0.32 & 0.68 & 98.45 & 99.32 & 99.67 & 87.35 & 73.56 & 52.89 & 91.23 \\ 
        \textbf{CVAE (}$\heartsuit$\textbf{)} & \textbf{0.23} & 38.27 & \textbf{0.12} & \textbf{0.34} & \textbf{0.72} & \textbf{99.33} & \textbf{99.67} & \textbf{99.84} & \textbf{87.73} & \textbf{73.97} & \textbf{53.45} & 91.18 \\ \hline
        TRS-Orig & 0.46 & 48.98 & 0.05 & 0.16 & 0.27 & \textbf{99.03} & 99.58 & 99.72 & 87.57 & 73.81 & 52.67 & 91.07 \\
        TRS-Ent & 0.47 & 54.10 & 0.10 & 0.45 & 0.94 & 97.26 & 98.97 & 99.51 & \textbf{87.69} & 73.99 & 50.88 & 91.24 \\
        \textbf{TRS (}$\spadesuit$\textbf{)} & 0.63 & 46.13 & 0.14 & 0.66 & 1.50 & 95.25 & 97.88 & 99.21 & 86.68 & 73.52 & 51.44 & 91.14 \\ 
        \textbf{TRS (}$\heartsuit$\textbf{)} & \textbf{0.80} & \textbf{46.06} & \textbf{0.17} & \textbf{0.78} & \textbf{1.70} & 98.57 & \textbf{99.61} & \textbf{99.76} & 86.00 & \textbf{74.40} & \textbf{53.57} & \textbf{91.55} \\ \hline
        GVT-Orig & 0.32 & \textbf{26.70} & 0.11 & 0.95 & 3.74 & 98.50 & 99.55 & 99.84 & 86.82 & 74.04 & 53.03 & 90.89 \\ 
        GVT-Ent & 0.32 & 30.76 & 0.08 & 0.65 & 2.48 & \textbf{98.62} & 99.56 & 99.81 & \textbf{87.32} & 73.07 & 50.70 & 91.52 \\
        \textbf{GVT (}$\spadesuit$\textbf{)} & 0.33 & 30.46 & 0.10 & 0.73 & 3.56 & 98.42 & 99.34 & 99.62 & 87.03 & 74.12 & 53.28 & 91.04 \\ 
        \textbf{GVT (}$\heartsuit$\textbf{)} & \textbf{0.35} & 29.78 & \textbf{0.16} & \textbf{1.06} & \textbf{4.27} & 98.55 & \textbf{99.63} & \textbf{99.91} & 87.25 & \textbf{75.10}& \textbf{54.12} & \textbf{91.63} \\
    \bottomrule[1pt]
    \end{tabular}}
    \label{tab:automatic1}
\end{table*}

\begin{table*}[!t]
    \centering
    \caption{Automatic evaluation results (\%) on the test set of DailyDialog.}
    \vspace{-2mm}
    \resizebox{\textwidth}{34mm}{
    \begin{tabular}{lcccccccccccc}
    \toprule[1pt]
        Model & BLEU & Perplexity$\downarrow$ & Dist-1 & Dist-2 & Dist-3 & Intra-1 & Intra-2 & Intra-3 & Average & Greedy & Extrema & Coherence \\
    \midrule[1pt]
        S2S-Orig & 1.13 & \textbf{33.43} & \textbf{1.69} & 6.10 & 11.04 & 94.22 & 97.29 & 98.15 & 90.33 & 76.73 & \textbf{52.50} & 89.37 \\
        S2S-Ent & \textbf{1.49} & 34.75 & 1.20 & 4.44 & 8.08 & 93.64 & 96.70 & 97.52 & \textbf{90.72} & 76.25 & 51.52 & \textbf{90.22} \\
        \textbf{S2S (}$\spadesuit$\textbf{)} & 1.35 & 34.04 & 1.57 & 5.79 & 10.27 & 95.84 & 97.92 & 98.83 & 90.31 & 76.43 & 52.37 & 89.45 \\ 
        \textbf{S2S (}$\heartsuit$\textbf{)} & 1.22 & 33.74 & 1.66 & \textbf{6.18} & \textbf{11.31} & \textbf{96.19} & \textbf{98.50} & \textbf{99.04} & 90.27 & \textbf{76.77} & 52.46 & 89.38 \\ \hline
        CVAE-Orig & 1.23 & \textbf{29.36} & 2.77 & 10.53 & 19.98 & 94.64 & 98.00 & 98.59 & 90.74 & 76.15 & 51.08 & 90.38 \\
        CVAE-Ent & \textbf{1.28} & 32.40 & 2.66 & 9.60 & 17.74 & 94.28 & 97.86 & 98.57 & 90.67 & 76.00 & \textbf{51.13} & \textbf{90.55} \\
        \textbf{CVAE (}$\spadesuit$\textbf{)} & 1.25 & 30.15 & 2.73 & 10.55 & 19.87 & 94.66 & \textbf{98.17} & 98.61 & 91.02 & 76.33 & 51.10 & 90.43 \\ 
        \textbf{CVAE (}$\heartsuit$\textbf{)} & 1.26 & 29.43 & \textbf{2.81} & \textbf{10.57} & \textbf{20.13} & \textbf{94.71} & 98.13 & \textbf{98.62} & \textbf{91.35} & \textbf{76.54} & 51.12 & 90.47 \\ \hline
        TRS-Orig & 2.42 & 31.13 & 1.21 & 4.62 & 9.22 & 83.10 & 91.77 & 94.25 & 91.04 & 76.58 & \textbf{50.59} & \textbf{91.50} \\ 
        TRS-Ent & \textbf{2.49} & 31.79 & 0.89 & 3.48 & 6.90 & 79.85 & 89.26 & 92.46 & 91.20 & 76.58 & 49.73 & 91.43 \\
        \textbf{TRS (}$\spadesuit$\textbf{)} & 2.40 & 31.70 & 0.93 & 4.20 & 8.07 & 83.21 & 91.29 & 93.63 & 91.26 & 76.51 & 49.90 & 91.11 \\ 
        \textbf{TRS (}$\heartsuit$\textbf{)} & 2.44 & \textbf{31.01} & \textbf{1.25} & \textbf{5.07} & \textbf{10.32} & \textbf{83.31} & \textbf{92.12} & \textbf{94.75} & \textbf{91.43} & \textbf{76.85} & 50.14 & 91.15 \\ \hline
        GVT-Orig & 1.14 & \textbf{30.04} & 0.30 & 1.37 & 3.76 & 92.22 & 97.09 & 98.06 & \textbf{91.63} & 76.84 & 50.33 & 91.30 \\
        GVT-Ent & 1.29 & 30.63 & 0.26 & 1.23 & 3.23 & 90.71 & 95.48 & 96.59 & 91.56 & 76.41 & 50.28 & \textbf{91.36} \\
        \textbf{GVT (}$\spadesuit$\textbf{)} & 1.25 & 30.25 & 0.29 & 1.34 & 3.72 & 92.08 & 96.45 & 97.86 & 91.52 & 76.53 & 50.34 & 91.34 \\ 
        \textbf{GVT (}$\heartsuit$\textbf{)} & \textbf{1.32} & 30.13 & \textbf{0.36} & \textbf{1.45} & \textbf{3.84} & \textbf{92.56} & \textbf{97.43} & \textbf{98.21} & 91.48 & \textbf{77.43} & \textbf{50.62} & 91.31 \\
    \bottomrule[1pt]
    \end{tabular}}
    \label{tab:automatic2}
\end{table*}

\subsection{Evaluation Measures}
We use both automatic metrics and human judgement for evaluation in our experiments.

\noindent\textbf{Automatic Metrics.}
We adopt several automatic metrics in existing work to measure the performance of dialogue generation models, including: (1) \textbf{BLEU} \cite{papineni2002bleu} measures how much a generated response containing n-gram overlaps with the ground-truth response\footnote{We use the script provided by \citet{lin2020variational} to calculate the BLEU score.}. (2) \textbf{Perplexity} \cite{serban2015hierarchical} measures the high-level general quality of the generation model, and usually a relatively lower Perplexity value indicates a more fluent response. (3) \textbf{Distinct} \cite{li2016diversity} measures the degree of diversity. Specifically, we leverage Dist-1/2/3 \cite{gu2018dialogwae} to compute the average of distinct values within each generated response. (4) \textbf{Embedding-based metrics} \cite{serban2017hierarchical,liu2016not,xu2018better} contain three metrics to compute the word embedding similarity between the generated response and the ground truth \cite{liu2016not}: 1. Greedy: greedily matching words in two utterances based on the cosine similarities between their embeddings; 2. Average: cosine similarity between the averaged word embeddings in the two utterances; 3. Extrema: cosine similarity between the largest extreme values among the word embeddings in the two utterances\footnote{We employ a popular NLG evaluation project available at \url{https://github.com/Maluuba/nlg-eval} for automatic evaluation.}. 
Inspired by \citet{xu2018better}, we use Coherence to calculate the cosine distance of two semantic vectors of a context and the generated response\footnote{The definition of Coherence in \citet{xu2018better} is similar to that of Relatedness in Section \ref{sec:attributes}.}. 
We use GloVe vectors pre-trained on Twitter as the word embeddings \cite{pennington2014glove}.

\noindent\textbf{Human Evaluation.}
A human evaluation is also conducted to validate the effectiveness of our proposed method. Firstly, we randomly sample 100 contexts from the test set and get the generated responses of NCMs trained on either filtered dataset or original dataset. Next, we send pairs of the context and generated response to three professional annotators without orders. Annotators are then required to evaluate among ``Win'' (response$_1$ is better), ``Loss'' (response$_2$ is better) and ``Tie'' (they are equally good or bad) independently, considering four aspects: Relatedness, Consistency, Fluency and Informativeness. Relatedness evaluates whether the generated responses are relevant on topic with its contexts; Consistency measures whether the generated responses contain factual contradictions with respect to its contexts; Fluency assesses the grammatical correctness and readability of the generated responses; Informativeness indicates whether the generated responses are informative and not generic. The result on each sample is determined by majority voting. Finally, we calculate the percentage of samples where an NCM trained on filtered or original dataset generates the better response and where an NCM performs similarly on filtered and original datasets.

\section{Experimental Results}
In this section, we demonstrate our experimental results on two datasets: OpenSubtitles and DailyDialog, including automatic evaluation results, human evaluation results, and case study on data filtering and dialogue generation.

\begin{table*}[h]
\centering
\caption{Results of human evaluation on two datasets: (1) OpenSubtitles and (2) DailyDialog. Tests are conducted pairwise between the responses generated by NCMs that are trained on filtered and original datasets. Agreements to measure the consistency among three annotators are calculated with the Fleiss' kappa \cite{fleiss1973equivalence}.}
\vspace{-2mm}
\begin{tabular}{l|lcccc|l|lcccc}
  \toprule[1pt] 
  & \textbf{Filtered vs. Original} & Win & Loss & Tie & Kappa & & \textbf{Filtered vs. Original} & Win & Loss & Tie & Kappa \\
  \midrule[1pt]
  \multirow{4}*{(1)} & S2S & 43\% & 15\% & 42\% & 0.514 & \multirow{4}*{(2)} & S2S & 43\% & 11\% & 46\% & 0.462 \\
  & CVAE & 48\% & 7\% & 45\% & 0.493 & & CVAE & 51\% & 9\% & 40\% & 0.584 \\
  & TRS & 52\% & 8\% & 40\% & 0.596 & & TRS & 45\% & 12\% & 43\% & 0.497 \\
  & GVT & 47\% & 10\% & 43\% & 0.565 & & GVT & 56\% & 8\% & 36\% & 0.515 \\
  \bottomrule[1pt]
\end{tabular}
\label{tab:human}
\end{table*}

\begin{table*}[!t]
\centering
\caption{Case study on Data Filtering on two datasets: (1) OpenSubtitles and (2) DailyDialog. ``$S$'' is the value calculated by Equation \ref{eq:S} to score each training sample.}
\vspace{-2mm}
\resizebox{0.93\textwidth}{20.5mm}{
\begin{tabular}{c|lllc}
  \toprule[1pt]
  & & Context & Response & $S$ \\
  \midrule[1pt]
  \multirow{4}*{(1)} & 1 & Get on with it. & No, No, No, No, No, No, No, No. & -1.2063 \\
  & 2 & Does he roll? $\rightarrow$ Does he talk in his sleep? $\rightarrow$ I don't know. & There's a spider! & 0.0010\\
  & 3 & Well, what do you want me to do? & Go back to harrow alley. Talk to Mrs.Goldstein. & 0.0748 \\
  & 4 & What? $\rightarrow$ The favorite flower for your bouquet. $\rightarrow$ Oh, well! & Red roses. & 1.6667 \\
  \midrule[1pt]
  \multirow{4}*{(2)} & 5 & You can get it from the student business office. & I didn't know that. & -0.6359 \\
  & 6 & That's all right. I understand. & I'm so sorry. & -0.4400 \\
  & 7 & I'm feeling hungry. & Isn't it time for dinner? & 0.7346 \\
  & 8 & How many rooms are there? & Eight rooms, ..., three sweet bedrooms. & 0.9264
 \\
  \bottomrule[1pt]
\end{tabular}}
\label{tab:scorecases}
\end{table*}

\begin{table*}[!t]
\centering
\caption{Case Study on Dialogue Generation with TRS on two datasets: (1) OpenSubtitles and (2) DailyDialog. ``-O'' and ``-F'' denote responses generated by a TRS model that is trained on the original datasets and filtered datasets, respectively.}
\vspace{-2mm}
\resizebox{0.93\textwidth}{20.5mm}{
\begin{tabular}{c|llll}
  \toprule[1pt]
  & & Context & Response-O & Response-F \\
  \midrule[1pt]
  \multirow{4}*{(1)} & 1 & Where were you? $\rightarrow$ On the toilet. &  I'm not going to get it. & I was just trying to get you. \\
  & 2 & She's in shock! Okay? & She's a good idea. & I'm sorry about that.\\
  & 3 & They're changing the guard over. & They 're not going to be a good idea. & Then they're not going to be here. \\
  & 4 & I totally need my board. I gotta go find it. &  I'm not going to get it. & OK, I'll see you later. \\
  \midrule[1pt]
  \multirow{4}*{(2)} & 5 & So, Paula, where are you from? & I'm in the middle of the city. & I'm from Beijing. \\
  & 6 & Hey man, you wanna buy some weed? & I' m sorry. I don't know what to do. & I' m sorry. I'm going to buy a new car. \\
  & 7 & ... I have a terrible headache. & I'm sorry, Sir. & I'm sorry to hear that. \\
  & 8 & Excuse me, may I help you ? & Yes, I'd like to buy a new one. & Yes, I'd like to book a book. \\
  \bottomrule[1pt]
\end{tabular}}
\label{tab:casestudy}
\end{table*}

\subsection{Automatic Evaluation Results (RQ1)} 
We list the results of automatic evaluation on both the OpenSubtitles and DailyDialog datasets in Table \ref{tab:automatic1} and \ref{tab:automatic2}, respectively. From the results, we have three main observations. 

First, compared with training on original datasets, our data filtering method can not only bring improvements for all the four NCMs on almost all the evaluation metrics (t-test, p-value $<$ 0.05), but also achieve competitive performance across two datasets. This also affirms the generalization ability of our proposed method. 

Second, when it comes to the comparison between our method and a comparable data filtering approach (``-Ent'') \cite{csaky2019improving}, our method can also achieve better performance. The work of \citet{csaky2019improving} aims to solve the ``general response'' problem and generate more diverse responses in open-domain conversations. Therefore, the higher Diversity scores, i.e., Dist-1/2/3 and Intra-1/2/3, further reflect the effectiveness of our proposed method.

Third, we also notice that the absolute improvement of Dist-1/2/3 value on OpenSubtitles is large (e.g., up to 0.12\%/0.62\%/1.32\% based on the TRS model), but for DailyDialog, the improvement is relatively small. We conjecture that OpenSubtitles benefits more from the data filtering method than DailyDialog, as DailyDialog is manually-collected and many samples of it are in high quality, while OpenSubtitles is more complex and contains a larger amount of low-quality samples.

\subsection{Human Evaluation Results (RQ1)}
We also conduct pairwise human evaluation to confirm the improvement of our method, and the results on two datasets are shown in Table \ref{tab:human}.  

We observe that training on filtered datasets outperforms training on original datasets for all the four NCMs as the percentage of ``Win'' is much larger than that of ``Loss''. The kappa scores indicate that the annotators came to a moderate agreement in the judgment. Meanwhile, as the DailyDialog dataset contains less low-quality training samples, the percentage of ``Tie'' is also high, showing that training on original or filtered dataset has similar performance. The human evaluation here is based on the overall sample quality, and we leave the fine-grained comparisons for the future work.

\begin{table*}[!t]
    \centering
    \caption{Kendall $\tau$ correlations among dialogue attributes on OpenSubtitles. This table illustrates that these attributes, in general, do not show strong correlations with each other.}
    \vspace{-2mm}
    \begin{tabular}{lc|lc|lc}
    \toprule[1pt]
        Opponent & Kendall & Opponent & Kendall & Opponent & Kendall \\ 
    \midrule[1pt]
        Specificity vs. Repetitiveness & 0.039 & Specificity vs. Relatedness & -0.044 & Specificity vs. Continuity & -0.057 \\ 
        Specificity vs. Coherence & -0.333 & 
        Specificity vs. Fluency & -0.451 &
        Specificity vs. Consistency & 0.005 \\
        Repetitiveness vs. Relatedness & 0.011 & Repetitiveness vs. Continuity & 0.011 & Repetitiveness vs. Coherence & -0.274 \\
        Repetitiveness vs. Fluency & 0.011 & Repetitiveness vs. Consistency & 0.025 & Relatedness vs. Continuity & 0.102 \\
        Relatedness vs. Coherence & 0.104 &
        Relatedness vs. Fluency & 0.005 &
        Relatedness vs. Consistency & 0.090 \\
        Continuity vs. Coherence & 0.047 &
        Continuity vs. Fluency & 0.027 &
        Continuity vs. Consistency & 0.035 \\
        Coherence vs. Fluency & 0.281 &
        Coherence vs. Consistency & -0.002 &
        Fluency vs. Consistency & 0.009 \\
    \bottomrule[1pt]
    \end{tabular}
    \label{tab:correlations}
\end{table*}

\subsection{Case Study on Data Filtering (RQ2)}
To better understand whether our method can identify untrustworthy samples, and even sort training samples based on data quality, we conduct some case studies on both OpenSubtitles and DailyDialog datasets. 

As shown in Table \ref{tab:scorecases}, we can see that cases with repetitive words, unrelated contents, or generic expressions, named as untrustworthy samples in our work, are scored with low $S$ scores. In contrast, informative and related cases are likely to have high $S$ scores. The results here are correlated with human judgement, which indicates the effectiveness of our designed measure $S$ that takes several attributes into consideration to assess the quality of a training sample.

\begin{table*}[!t]
    \centering
    \caption{Ablation Study on OpenSubtitles with TRS. ``None'' and ``$S$'' denote training on original dataset and filtered dataset with our proposed method, respectively. Others lines represent training on the dataset filtered by a single dialogue attribute. Average, Greedy, Extrema, and Coherence are abbreviated as Avg., Gre., Ext., and Coh., respectively. This table shows that data filtering can lead to better performance, even with only one attribute. When applying the measure $S$ that combines seven attributes, the performance is the best on most of automatic metrics.}
    \vspace{-2mm}
    \resizebox{0.94\textwidth}{24mm}{
    \begin{tabular}{l|cccccccccccc}
    \toprule[1pt]
        Filtered by & BLEU & Perplexity$\downarrow$ & Dist-1 & Dist-2 & Dist-3 & Intra-1 & Intra-2 & Intra-3 & Avg. & Gre. & Ext. & Coh. \\ 
    \midrule[1pt]
        None & 0.46 & 48.98 & 0.05 & 0.16 & 0.27 & 99.03 & 99.58 & 99.72 & 87.57 & 73.81 & 52.67 & 91.07 \\ 
    \midrule[1pt]
        Specificity & 0.42 & 51.65 & 0.05 & 0.15 & 0.27 & 98.33 & 99.31 & 98.91 & \textbf{87.97} & 73.84 & 50.98 & 91.50 \\
        Repetitiveness & 0.61 & 47.24 & 0.09 & 0.34 & 0.63 & \textbf{99.26} & \textbf{99.76} & \textbf{99.86} & 86.00 & 72.30 & 50.47 & 90.28 \\
        Relatedness & 0.55 & 47.48 & 0.08 & 0.27 & 0.51 & 92.15 & 94.81 & 97.16 & 86.01 & 72.41 & 51.21 & 90.37 \\
        Continuity & \textbf{0.83} & 47.12 & 0.09 & 0.36 & 0.69 & 97.80 & 98.55 & 98.90 & 87.34 & 74.15 & 53.31 & 91.34 \\
        Coherence & 0.65 & 47.07 & 0.11 & 0.46 & 0.93 & 97.79 & 98.91 & 99.35 & 86.58 & 73.29 & 51.92 & 91.37 \\
        Fluency & 0.68 & 47.16 & 0.12 & 0.51 & 1.08 & 98.16 & 98.81 & 99.01 & 85.34 & 73.01 & 52.97 & 89.70 \\
        Consistency & 0.71 & 47.18 & 0.11 & 0.47 & 0.93 & 97.77 & 98.69 & 98.92 & 86.33 & 72.79 & 50.90 & 91.20 \\ 
    \midrule[1pt]
        $S$ & 0.80 & \textbf{46.06} & \textbf{0.17} & \textbf{0.78} & \textbf{1.70} & 98.57 & 99.61 & 99.76 & 86.00 & \textbf{74.40} & \textbf{53.57} & \textbf{91.55} \\
    \bottomrule[1pt]
    \end{tabular}}
    \label{tab:ablation}
\end{table*}

\subsection{Case Study on Dialogue Generation (RQ3)}
The ultimate goal of our work is to improve the open-domain dialogue generation with data filtering method. To further check the performance of NCMs trained on filtered datasets, we show some responses generated by the TRS model on both OpenSubtitles and DailyDialog datasets in Table \ref{tab:casestudy}.

``Response-O'' and ``Response-F'' denote responses generated by a TRS model that is trained on original datasets and filtered datasets, respectively. In general, responses under ``Response-F'' are more informative and related to the corresponding contexts. Take Case 1 as an example. The context contains a question ``Where were you?'', and the response is ``I was just trying to get you''. The past tenses and contents in these two utterances make them highly coherent with each other. However, ``I'm not going to get it'' is irrelevant to the context, especially with a strange ``it'' that confuses people a lot.

\section{Further Analysis}
We conduct some further analyses to validate the effectiveness of our method.

\subsection{Correlation among Attributes (RQ4)}
Seven dialogue attributes are combined together when we define $\varphi(X)$ in Section \ref{sec:attributes}. To show their relations, we calculate the Kendall $\tau$ correlations of any two attributes inspired by \citet{cai2020learning}. Table \ref{tab:correlations} illustrates the results on OpenSubtitles. 

We see that the maximum absolute value of correlations is 0.451 between Specificity and Fluency, and other correlation values are around 0.1. This demonstrates that these attributes, in general, do not show strong correlations with each other. At the same time, it partially validates that dialogue quality is reflected in multiple facets. Besides, the negative correlation of Specificity and Fluency is also consistent with people's cognition, as a generic response has a higher probability to be fluent.

\subsection{Single vs. Multiple Attributes (RQ5)}
To gain some insights into the effects of seven dialogue attributes on our proposed filtering method, we conduct the ablation study using TRS by only exploiting a single attribute as the measure to score each training sample. Table \ref{tab:ablation} reports the results on Opensubtitles. 

We observe that data filtering can lead to better performance, even with only one attribute. When applying the measure $S$ that combines seven attributes, the performance is the best on most of automatic metrics, including Perplexity, Diversity metrics, and Embedding-based metrics representing fluency, diversity, and relatedness, respectively,

\subsection{$J$-value Curves of Optimization (RQ6)}
\begin{figure}[t]
\begin{center}
   \includegraphics[width=0.95\linewidth]{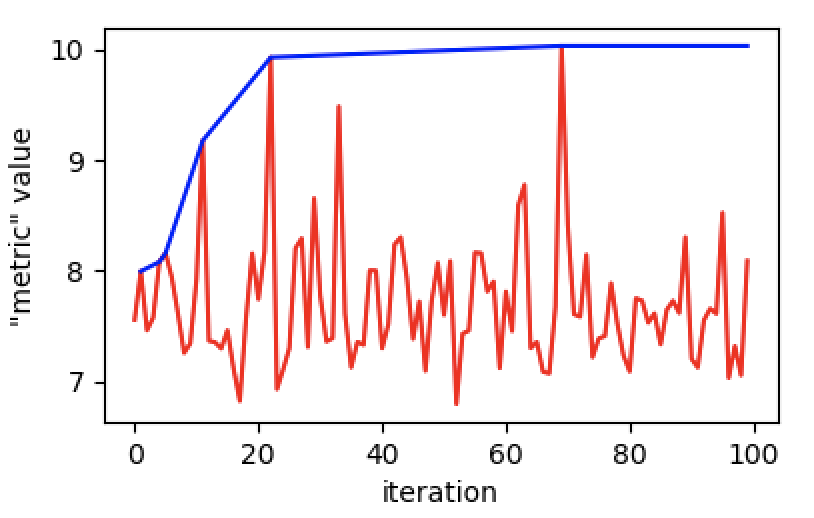}
\end{center}
\vspace{-2mm}
   \caption{$J$-value curves of Bayesian Optimization on the validation set. The red line denotes the hypothesis space exploration of Bayesian Optimization for the ``metric'' value, while the blue line displays the overall best value on the validation set. We find that Bayesian Optimization here results in a large explored space (more variance), and the ``metric'' value keeps increasing as the iteration continues.}
\label{fig:J}
\end{figure}

To get some intuitive feelings of the optimization procedure, Figure \ref{fig:J} shows the ``metric'' value, the sum of 13 open-domain automatic metrics, for model TRS on OpenSubtitles. The red line denotes the hypothesis space exploration of Bayesian Optimization for the ``metric'' value\footnote{Bayesian Optimization aims to minimize ``-metric'' value, while here we plot the increase of ``metric'' value.}, while the blue line displays the overall best value on the validation set. 

We find that Bayesian Optimization here results in a large explored space (more variance), and the ``metric'' value keeps increasing as the iteration continues. This plot can not only illustrate the optimization produce, but also implicitly validate the effectiveness of DIFF-MLE-NEG, as Bayesian Optimization can work normally.

\section{Related Work}
Our work is related to two research branches: Data Filtering for Noisy Corpora, and Metrics for Automatic Dialogue Evaluation.
\label{relatedwork}

\subsection{Data Filtering for Noisy Corpora.}
\label{sec:related3}
To better utilize noisy dialogue corpora, model- and data-based approaches have been proposed. The former one use models to handle noise during the training process. \citet{shang2018learning} integrated a calibration network into a generation network. The calibration network is trained to measure the quality of the context-response pairs, and the generation network takes the scores produced by the calibration network to weight the training samples, such that the high-quality samples have more impacts on the generation model while the low-quality ones are less influential. The latter one aims to improve the data quality by pre-processing before feeding it into a model. \citet{xu2018better} introduced \textit{Coherence} with cosine similarity to measure the semantic relationship between contexts and responses, and then filtered out pairs with low coherence score. \citet{csaky2019improving} used entropy to find out the most generic contexts or responses, and then filtered out pairs with high entropy. \citet{akama2020filtering} defined \textit{Connectivity (C)} and \textit{Relatedness (R)} with normalized pointwise mutual information \cite{bouma2009normalized} and cosine similarity to remove noisy pairs with low \textit{C+R} score. \citet{li2017data} proposed an iteration-based way to distill data, which operates as: a neural sequence-to-sequence (S2S) model is first trained and used to generate responses to inputs in a dataset. Then a list of the most common responses is constructed, and training samples with outputs that are semantically close to these common responses are removed. As the process iterates, responses that are generic are gradually distilled, and the trained models gradually increase in specificity.

The differences between our method and the above listed ones are: (1) The proposed measure $S$ does not rely on only one or two attributes, it integrates seven dialogue attributes to estimate the quality of dialogue samples. (2) We frame the dialogue data filtering task as an optimization problem, and use Bayesian Optimization to find the optimal weights in the linear combination of $S$.

\subsection{Metrics for Automatic Dialogue Evaluation.}
\label{sec:related2}
The evaluation of open-domain dialogue generation generally consists of both automatic metrics and human judgement. Since human evaluation has high variance, high cost, and is difficult to replicate, some researchers aim to propose automatic metrics that correlate more with human intuition for this task. Dialogue quality is inherently multi-faceted \cite{walker1997paradise,see2019makes,cai2020learning}, including fluency, coherence, diversity, consistency, etc. Inspired by the automatic evaluation conducted in other fields, referenced automatic metrics are applied \cite{papineni2002bleu,ghazarian2019better,tao2018ruber}. Recently, unreferenced metrics are also proposed to assess different dialogue attributes, such as Distinct \cite{li2016diversity}, Entropy \cite{pang2020towards}, MAUDE \cite{sinha2020learning}, USR \cite{mehri2020usr}, and FED \cite{mehri2020unsupervised}. The last three ones also rely on pre-trained language models, and have higher correlation with human judgement. However, compared with some simple metrics, these new metrics are not widely used and thoroughly accepted by this community.

Our work does \textbf{not} aim to propose a new automatic metric for open-domain dialogue generation, we only utilize some of them to help define the measure $S$ or objective function $J$. In addition to the options we choose in Section \ref{sec:attributes}, other metrics like FED can also be used to compute $S$ or $J$, and we leave this for the future work.

\section{Conclusion and Future Work}
\label{conclusion}
In this paper, we present a data filtering method for open-domain dialogues, which aims to identify untrustworthy training samples with a quality measure that linearly combines seven dialogue attributes. The attribute weights are obtained via Bayesian Optimization. Besides, to accelerate the optimization iterations on large-scale datasets, we propose a training method that integrates the maximum likelihood estimation (MLE) and negative training method (NEG). Experimental results on two widely-used datasets show that our method can effectively identify untrustworthy samples, and NCMs trained on the filtered datasets can further generate fluent, related, and informative responses.
In the future, we will try to utilize some newly proposed automatic metrics for open-domain dialogue generation, and combine them in the definition of the quality measure and the choice of the objective function. 

\bibliographystyle{ACM-Reference-Format}
\balance
\bibliography{cikm2021ref}


\begin{thebibliography}{61}


\ifx \showCODEN    \undefined \def \showCODEN     #1{\unskip}     \fi
\ifx \showDOI      \undefined \def \showDOI       #1{#1}\fi
\ifx \showISBNx    \undefined \def \showISBNx     #1{\unskip}     \fi
\ifx \showISBNxiii \undefined \def \showISBNxiii  #1{\unskip}     \fi
\ifx \showISSN     \undefined \def \showISSN      #1{\unskip}     \fi
\ifx \showLCCN     \undefined \def \showLCCN      #1{\unskip}     \fi
\ifx \shownote     \undefined \def \shownote      #1{#1}          \fi
\ifx \showarticletitle \undefined \def \showarticletitle #1{#1}   \fi
\ifx \showURL      \undefined \def \showURL       {\relax}        \fi
\providecommand\bibfield[2]{#2}
\providecommand\bibinfo[2]{#2}
\providecommand\natexlab[1]{#1}
\providecommand\showeprint[2][]{arXiv:#2}

\bibitem[\protect\citeauthoryear{Akama, Yokoi, Suzuki, and Inui}{Akama
  et~al\mbox{.}}{2020}]%
        {akama2020filtering}
\bibfield{author}{\bibinfo{person}{Reina Akama}, \bibinfo{person}{Sho Yokoi},
  \bibinfo{person}{Jun Suzuki}, {and} \bibinfo{person}{Kentaro Inui}.}
  \bibinfo{year}{2020}\natexlab{}.
\newblock \showarticletitle{Filtering Noisy Dialogue Corpora by Connectivity
  and Content Relatedness}. In \bibinfo{booktitle}{\emph{Proceedings of the
  2020 Conference on Empirical Methods in Natural Language Processing
  (EMNLP)}}. \bibinfo{publisher}{Association for Computational Linguistics},
  \bibinfo{address}{Online}, \bibinfo{pages}{941--958}.
\newblock


\bibitem[\protect\citeauthoryear{Arora, Liang, and Ma}{Arora
  et~al\mbox{.}}{2017}]%
        {arora2016simple}
\bibfield{author}{\bibinfo{person}{Sanjeev Arora}, \bibinfo{person}{Yingyu
  Liang}, {and} \bibinfo{person}{Tengyu Ma}.} \bibinfo{year}{2017}\natexlab{}.
\newblock \showarticletitle{A Simple but Tough-to-Beat Baseline for Sentence
  Embeddings}. In \bibinfo{booktitle}{\emph{5th International Conference on
  Learning Representations, {ICLR} 2017, Toulon, France, April 24-26, 2017,
  Conference Track Proceedings}}. \bibinfo{publisher}{OpenReview.net}.
\newblock


\bibitem[\protect\citeauthoryear{Bahdanau, Cho, and Bengio}{Bahdanau
  et~al\mbox{.}}{2015}]%
        {bahdanau2014neural}
\bibfield{author}{\bibinfo{person}{Dzmitry Bahdanau},
  \bibinfo{person}{Kyunghyun Cho}, {and} \bibinfo{person}{Yoshua Bengio}.}
  \bibinfo{year}{2015}\natexlab{}.
\newblock \showarticletitle{Neural Machine Translation by Jointly Learning to
  Align and Translate}. In \bibinfo{booktitle}{\emph{3rd International
  Conference on Learning Representations, {ICLR} 2015, San Diego, CA, USA, May
  7-9, 2015, Conference Track Proceedings}},
  \bibfield{editor}{\bibinfo{person}{Yoshua Bengio} {and} \bibinfo{person}{Yann
  LeCun}} (Eds.).
\newblock


\bibitem[\protect\citeauthoryear{Bouma}{Bouma}{2009}]%
        {bouma2009normalized}
\bibfield{author}{\bibinfo{person}{Gerlof Bouma}.}
  \bibinfo{year}{2009}\natexlab{}.
\newblock \showarticletitle{Normalized (pointwise) mutual information in
  collocation extraction}.
\newblock  (\bibinfo{year}{2009}).
\newblock


\bibitem[\protect\citeauthoryear{Bowman, Angeli, Potts, and Manning}{Bowman
  et~al\mbox{.}}{2015}]%
        {bowman2015large}
\bibfield{author}{\bibinfo{person}{Samuel~R. Bowman}, \bibinfo{person}{Gabor
  Angeli}, \bibinfo{person}{Christopher Potts}, {and}
  \bibinfo{person}{Christopher~D. Manning}.} \bibinfo{year}{2015}\natexlab{}.
\newblock \showarticletitle{A large annotated corpus for learning natural
  language inference}. In \bibinfo{booktitle}{\emph{Proceedings of the 2015
  Conference on Empirical Methods in Natural Language Processing}}.
  \bibinfo{publisher}{Association for Computational Linguistics},
  \bibinfo{address}{Lisbon, Portugal}, \bibinfo{pages}{632--642}.
\newblock


\bibitem[\protect\citeauthoryear{Brochu, Cora, and De~Freitas}{Brochu
  et~al\mbox{.}}{2010}]%
        {brochu2010tutorial}
\bibfield{author}{\bibinfo{person}{Eric Brochu}, \bibinfo{person}{Vlad~M Cora},
  {and} \bibinfo{person}{Nando De~Freitas}.} \bibinfo{year}{2010}\natexlab{}.
\newblock \showarticletitle{A tutorial on Bayesian optimization of expensive
  cost functions, with application to active user modeling and hierarchical
  reinforcement learning}.
\newblock \bibinfo{journal}{\emph{arXiv preprint arXiv:1012.2599}}
  (\bibinfo{year}{2010}).
\newblock


\bibitem[\protect\citeauthoryear{Cai, Chen, Zhang, Song, Zhao, Li, Duan, and
  Yin}{Cai et~al\mbox{.}}{2020}]%
        {cai2020learning}
\bibfield{author}{\bibinfo{person}{Hengyi Cai}, \bibinfo{person}{Hongshen
  Chen}, \bibinfo{person}{Cheng Zhang}, \bibinfo{person}{Yonghao Song},
  \bibinfo{person}{Xiaofang Zhao}, \bibinfo{person}{Yangxi Li},
  \bibinfo{person}{Dongsheng Duan}, {and} \bibinfo{person}{Dawei Yin}.}
  \bibinfo{year}{2020}\natexlab{}.
\newblock \showarticletitle{Learning from Easy to Complex: Adaptive
  Multi-Curricula Learning for Neural Dialogue Generation}. In
  \bibinfo{booktitle}{\emph{The Thirty-Fourth {AAAI} Conference on Artificial
  Intelligence, {AAAI} 2020, The Thirty-Second Innovative Applications of
  Artificial Intelligence Conference, {IAAI} 2020, The Tenth {AAAI} Symposium
  on Educational Advances in Artificial Intelligence, {EAAI} 2020, New York,
  NY, USA, February 7-12, 2020}}. \bibinfo{publisher}{{AAAI} Press},
  \bibinfo{pages}{7472--7479}.
\newblock


\bibitem[\protect\citeauthoryear{Chen, Liu, Shen, Yuan, Zhou, Wu, He, and
  Zhou}{Chen et~al\mbox{.}}{2020}]%
        {chen2020jddc}
\bibfield{author}{\bibinfo{person}{Meng Chen}, \bibinfo{person}{Ruixue Liu},
  \bibinfo{person}{Lei Shen}, \bibinfo{person}{Shaozu Yuan},
  \bibinfo{person}{Jingyan Zhou}, \bibinfo{person}{Youzheng Wu},
  \bibinfo{person}{Xiaodong He}, {and} \bibinfo{person}{Bowen Zhou}.}
  \bibinfo{year}{2020}\natexlab{}.
\newblock \showarticletitle{The {JDDC} Corpus: A Large-Scale Multi-Turn
  {C}hinese Dialogue Dataset for {E}-commerce Customer Service}. In
  \bibinfo{booktitle}{\emph{Proceedings of the 12th Language Resources and
  Evaluation Conference}}. \bibinfo{publisher}{European Language Resources
  Association}, \bibinfo{address}{Marseille, France},
  \bibinfo{pages}{459--466}.
\newblock


\bibitem[\protect\citeauthoryear{Cho, Gulcehre, Bahdanau, Schwenk, and
  Bengio}{Cho et~al\mbox{.}}{2014}]%
        {cholearning}
\bibfield{author}{\bibinfo{person}{Kyunghyun Cho}, \bibinfo{person}{Bart van
  Merri{\"e}nboer~Caglar Gulcehre}, \bibinfo{person}{Dzmitry Bahdanau},
  \bibinfo{person}{Fethi Bougares~Holger Schwenk}, {and}
  \bibinfo{person}{Yoshua Bengio}.} \bibinfo{year}{2014}\natexlab{}.
\newblock \showarticletitle{Learning Phrase Representations using RNN
  Encoder--Decoder for Statistical Machine Translation}. In
  \bibinfo{booktitle}{\emph{Proceedings of the 2014 Conference on Empirical
  Methods in Natural Language Processing}}. \bibinfo{pages}{1724--1734}.
\newblock


\bibitem[\protect\citeauthoryear{Cs{\'a}ky, Purgai, and Recski}{Cs{\'a}ky
  et~al\mbox{.}}{2019}]%
        {csaky2019improving}
\bibfield{author}{\bibinfo{person}{Rich{\'a}rd Cs{\'a}ky},
  \bibinfo{person}{Patrik Purgai}, {and} \bibinfo{person}{G{\'a}bor Recski}.}
  \bibinfo{year}{2019}\natexlab{}.
\newblock \showarticletitle{Improving Neural Conversational Models with
  Entropy-Based Data Filtering}. In \bibinfo{booktitle}{\emph{Proceedings of
  the 57th Annual Meeting of the Association for Computational Linguistics}}.
  \bibinfo{publisher}{Association for Computational Linguistics},
  \bibinfo{address}{Florence, Italy}, \bibinfo{pages}{5650--5669}.
\newblock


\bibitem[\protect\citeauthoryear{Fleiss and Cohen}{Fleiss and Cohen}{1973}]%
        {fleiss1973equivalence}
\bibfield{author}{\bibinfo{person}{Joseph~L Fleiss} {and}
  \bibinfo{person}{Jacob Cohen}.} \bibinfo{year}{1973}\natexlab{}.
\newblock \showarticletitle{The equivalence of weighted kappa and the
  intraclass correlation coefficient as measures of reliability}.
\newblock \bibinfo{journal}{\emph{Educational and psychological measurement}}
  \bibinfo{volume}{33}, \bibinfo{number}{3} (\bibinfo{year}{1973}),
  \bibinfo{pages}{613--619}.
\newblock


\bibitem[\protect\citeauthoryear{Ghazarian, Wei, Galstyan, and Peng}{Ghazarian
  et~al\mbox{.}}{2019}]%
        {ghazarian2019better}
\bibfield{author}{\bibinfo{person}{Sarik Ghazarian}, \bibinfo{person}{Johnny
  Tian-Zheng Wei}, \bibinfo{person}{Aram Galstyan}, {and}
  \bibinfo{person}{Nanyun Peng}.} \bibinfo{year}{2019}\natexlab{}.
\newblock \showarticletitle{Better Automatic Evaluation of Open-Domain Dialogue
  Systems with Contextualized Embeddings}.
\newblock \bibinfo{journal}{\emph{NAACL HLT 2019}} (\bibinfo{year}{2019}),
  \bibinfo{pages}{82}.
\newblock


\bibitem[\protect\citeauthoryear{Gu, Cho, Ha, and Kim}{Gu
  et~al\mbox{.}}{2019}]%
        {gu2018dialogwae}
\bibfield{author}{\bibinfo{person}{Xiaodong Gu}, \bibinfo{person}{Kyunghyun
  Cho}, \bibinfo{person}{Jung{-}Woo Ha}, {and} \bibinfo{person}{Sunghun Kim}.}
  \bibinfo{year}{2019}\natexlab{}.
\newblock \showarticletitle{DialogWAE: Multimodal Response Generation with
  Conditional Wasserstein Auto-Encoder}. In \bibinfo{booktitle}{\emph{7th
  International Conference on Learning Representations, {ICLR} 2019, New
  Orleans, LA, USA, May 6-9, 2019}}. \bibinfo{publisher}{OpenReview.net}.
\newblock


\bibitem[\protect\citeauthoryear{Gulrajani, Ahmed, Arjovsky, Dumoulin, and
  Courville}{Gulrajani et~al\mbox{.}}{2017}]%
        {gulrajani2017improved}
\bibfield{author}{\bibinfo{person}{Ishaan Gulrajani}, \bibinfo{person}{Faruk
  Ahmed}, \bibinfo{person}{Martin Arjovsky}, \bibinfo{person}{Vincent
  Dumoulin}, {and} \bibinfo{person}{Aaron Courville}.}
  \bibinfo{year}{2017}\natexlab{}.
\newblock \showarticletitle{Improved training of wasserstein GANs}. In
  \bibinfo{booktitle}{\emph{Proceedings of the 31st International Conference on
  Neural Information Processing Systems}}. \bibinfo{pages}{5769--5779}.
\newblock


\bibitem[\protect\citeauthoryear{He and Glass}{He and Glass}{2020}]%
        {he2019negative}
\bibfield{author}{\bibinfo{person}{Tianxing He} {and} \bibinfo{person}{James
  Glass}.} \bibinfo{year}{2020}\natexlab{}.
\newblock \showarticletitle{Negative Training for Neural Dialogue Response
  Generation}. In \bibinfo{booktitle}{\emph{Proceedings of the 58th Annual
  Meeting of the Association for Computational Linguistics}}.
  \bibinfo{publisher}{Association for Computational Linguistics},
  \bibinfo{address}{Online}, \bibinfo{pages}{2044--2058}.
\newblock


\bibitem[\protect\citeauthoryear{Henderson, Budzianowski, Casanueva, Coope,
  Gerz, Kumar, Mrk{\v{s}}i{\'c}, Spithourakis, Su, Vuli{\'c}, and
  Wen}{Henderson et~al\mbox{.}}{2019}]%
        {henderson2019repository}
\bibfield{author}{\bibinfo{person}{Matthew Henderson},
  \bibinfo{person}{Pawe{\l} Budzianowski}, \bibinfo{person}{I{\~n}igo
  Casanueva}, \bibinfo{person}{Sam Coope}, \bibinfo{person}{Daniela Gerz},
  \bibinfo{person}{Girish Kumar}, \bibinfo{person}{Nikola Mrk{\v{s}}i{\'c}},
  \bibinfo{person}{Georgios Spithourakis}, \bibinfo{person}{Pei-Hao Su},
  \bibinfo{person}{Ivan Vuli{\'c}}, {and} \bibinfo{person}{Tsung-Hsien Wen}.}
  \bibinfo{year}{2019}\natexlab{}.
\newblock \showarticletitle{A Repository of Conversational Datasets}. In
  \bibinfo{booktitle}{\emph{Proceedings of the First Workshop on NLP for
  Conversational AI}}. \bibinfo{publisher}{Association for Computational
  Linguistics}, \bibinfo{address}{Florence, Italy}, \bibinfo{pages}{1--10}.
\newblock


\bibitem[\protect\citeauthoryear{Ke, Guan, Huang, and Zhu}{Ke
  et~al\mbox{.}}{2018}]%
        {ke2018generating}
\bibfield{author}{\bibinfo{person}{Pei Ke}, \bibinfo{person}{Jian Guan},
  \bibinfo{person}{Minlie Huang}, {and} \bibinfo{person}{Xiaoyan Zhu}.}
  \bibinfo{year}{2018}\natexlab{}.
\newblock \showarticletitle{Generating Informative Responses with Controlled
  Sentence Function}. In \bibinfo{booktitle}{\emph{Proceedings of the 56th
  Annual Meeting of the Association for Computational Linguistics (Volume 1:
  Long Papers)}}. \bibinfo{publisher}{Association for Computational
  Linguistics}, \bibinfo{address}{Melbourne, Australia},
  \bibinfo{pages}{1499--1508}.
\newblock


\bibitem[\protect\citeauthoryear{Li, Galley, Brockett, Gao, and Dolan}{Li
  et~al\mbox{.}}{2016a}]%
        {li2016diversity}
\bibfield{author}{\bibinfo{person}{Jiwei Li}, \bibinfo{person}{Michel Galley},
  \bibinfo{person}{Chris Brockett}, \bibinfo{person}{Jianfeng Gao}, {and}
  \bibinfo{person}{Bill Dolan}.} \bibinfo{year}{2016}\natexlab{a}.
\newblock \showarticletitle{A Diversity-Promoting Objective Function for Neural
  Conversation Models}. In \bibinfo{booktitle}{\emph{Proceedings of the 2016
  Conference of the North {A}merican Chapter of the Association for
  Computational Linguistics: Human Language Technologies}}.
  \bibinfo{publisher}{Association for Computational Linguistics},
  \bibinfo{address}{San Diego, California}, \bibinfo{pages}{110--119}.
\newblock


\bibitem[\protect\citeauthoryear{Li, Galley, Brockett, Spithourakis, Gao, and
  Dolan}{Li et~al\mbox{.}}{2016b}]%
        {li2016persona}
\bibfield{author}{\bibinfo{person}{Jiwei Li}, \bibinfo{person}{Michel Galley},
  \bibinfo{person}{Chris Brockett}, \bibinfo{person}{Georgios Spithourakis},
  \bibinfo{person}{Jianfeng Gao}, {and} \bibinfo{person}{Bill Dolan}.}
  \bibinfo{year}{2016}\natexlab{b}.
\newblock \showarticletitle{A Persona-Based Neural Conversation Model}. In
  \bibinfo{booktitle}{\emph{Proceedings of the 54th Annual Meeting of the
  Association for Computational Linguistics (Volume 1: Long Papers)}}.
  \bibinfo{publisher}{Association for Computational Linguistics},
  \bibinfo{address}{Berlin, Germany}, \bibinfo{pages}{994--1003}.
\newblock


\bibitem[\protect\citeauthoryear{Li, Monroe, and Jurafsky}{Li
  et~al\mbox{.}}{2017a}]%
        {li2017data}
\bibfield{author}{\bibinfo{person}{Jiwei Li}, \bibinfo{person}{Will Monroe},
  {and} \bibinfo{person}{Dan Jurafsky}.} \bibinfo{year}{2017}\natexlab{a}.
\newblock \showarticletitle{Data distillation for controlling specificity in
  dialogue generation}.
\newblock \bibinfo{journal}{\emph{arXiv preprint arXiv:1702.06703}}
  (\bibinfo{year}{2017}).
\newblock


\bibitem[\protect\citeauthoryear{Li, Su, Shen, Li, Cao, and Niu}{Li
  et~al\mbox{.}}{2017b}]%
        {li2017dailydialog}
\bibfield{author}{\bibinfo{person}{Yanran Li}, \bibinfo{person}{Hui Su},
  \bibinfo{person}{Xiaoyu Shen}, \bibinfo{person}{Wenjie Li},
  \bibinfo{person}{Ziqiang Cao}, {and} \bibinfo{person}{Shuzi Niu}.}
  \bibinfo{year}{2017}\natexlab{b}.
\newblock \showarticletitle{{D}aily{D}ialog: A Manually Labelled Multi-turn
  Dialogue Dataset}. In \bibinfo{booktitle}{\emph{Proceedings of the Eighth
  International Joint Conference on Natural Language Processing (Volume 1: Long
  Papers)}}. \bibinfo{publisher}{Asian Federation of Natural Language
  Processing}, \bibinfo{address}{Taipei, Taiwan}, \bibinfo{pages}{986--995}.
\newblock


\bibitem[\protect\citeauthoryear{Lin, Winata, Xu, Liu, and Fung}{Lin
  et~al\mbox{.}}{2020}]%
        {lin2020variational}
\bibfield{author}{\bibinfo{person}{Zhaojiang Lin}, \bibinfo{person}{Genta~Indra
  Winata}, \bibinfo{person}{Peng Xu}, \bibinfo{person}{Zihan Liu}, {and}
  \bibinfo{person}{Pascale Fung}.} \bibinfo{year}{2020}\natexlab{}.
\newblock \showarticletitle{Variational transformers for diverse response
  generation}.
\newblock \bibinfo{journal}{\emph{arXiv preprint arXiv:2003.12738}}
  (\bibinfo{year}{2020}).
\newblock


\bibitem[\protect\citeauthoryear{Lison and Tiedemann}{Lison and
  Tiedemann}{2016}]%
        {lison2016opensubtitles2016}
\bibfield{author}{\bibinfo{person}{Pierre Lison} {and}
  \bibinfo{person}{J{\"o}rg Tiedemann}.} \bibinfo{year}{2016}\natexlab{}.
\newblock \showarticletitle{{O}pen{S}ubtitles2016: Extracting Large Parallel
  Corpora from Movie and {TV} Subtitles}. In
  \bibinfo{booktitle}{\emph{Proceedings of the Tenth International Conference
  on Language Resources and Evaluation ({LREC}'16)}}.
  \bibinfo{publisher}{European Language Resources Association (ELRA)},
  \bibinfo{address}{Portoro{\v{z}}, Slovenia}, \bibinfo{pages}{923--929}.
\newblock


\bibitem[\protect\citeauthoryear{Liu, Lowe, Serban, Noseworthy, Charlin, and
  Pineau}{Liu et~al\mbox{.}}{2016}]%
        {liu2016not}
\bibfield{author}{\bibinfo{person}{Chia-Wei Liu}, \bibinfo{person}{Ryan Lowe},
  \bibinfo{person}{Iulian Serban}, \bibinfo{person}{Mike Noseworthy},
  \bibinfo{person}{Laurent Charlin}, {and} \bibinfo{person}{Joelle Pineau}.}
  \bibinfo{year}{2016}\natexlab{}.
\newblock \showarticletitle{How {NOT} To Evaluate Your Dialogue System: An
  Empirical Study of Unsupervised Evaluation Metrics for Dialogue Response
  Generation}. In \bibinfo{booktitle}{\emph{Proceedings of the 2016 Conference
  on Empirical Methods in Natural Language Processing}}.
  \bibinfo{publisher}{Association for Computational Linguistics},
  \bibinfo{address}{Austin, Texas}, \bibinfo{pages}{2122--2132}.
\newblock


\bibitem[\protect\citeauthoryear{Liu, Chen, Liu, Shen, Song, and He}{Liu
  et~al\mbox{.}}{2020}]%
        {liu2020nlpcc}
\bibfield{author}{\bibinfo{person}{Ruixue Liu}, \bibinfo{person}{Meng Chen},
  \bibinfo{person}{Hang Liu}, \bibinfo{person}{Lei Shen}, \bibinfo{person}{Yang
  Song}, {and} \bibinfo{person}{Xiaodong He}.} \bibinfo{year}{2020}\natexlab{}.
\newblock \showarticletitle{Enhancing Multi-turn Dialogue Modeling with Intent
  Information for E-Commerce Customer Service}. In
  \bibinfo{booktitle}{\emph{CCF International Conference on Natural Language
  Processing and Chinese Computing}}. Springer, \bibinfo{pages}{65--77}.
\newblock


\bibitem[\protect\citeauthoryear{Liu, Ott, Goyal, Du, Joshi, Chen, Levy, Lewis,
  Zettlemoyer, and Stoyanov}{Liu et~al\mbox{.}}{2019b}]%
        {liu2019roberta}
\bibfield{author}{\bibinfo{person}{Yinhan Liu}, \bibinfo{person}{Myle Ott},
  \bibinfo{person}{Naman Goyal}, \bibinfo{person}{Jingfei Du},
  \bibinfo{person}{Mandar Joshi}, \bibinfo{person}{Danqi Chen},
  \bibinfo{person}{Omer Levy}, \bibinfo{person}{Mike Lewis},
  \bibinfo{person}{Luke Zettlemoyer}, {and} \bibinfo{person}{Veselin
  Stoyanov}.} \bibinfo{year}{2019}\natexlab{b}.
\newblock \showarticletitle{RoBERTa: A Robustly Optimized BERT Pretraining
  Approach}.
\newblock  (\bibinfo{year}{2019}).
\newblock


\bibitem[\protect\citeauthoryear{Liu, Niu, Wu, and Wang}{Liu
  et~al\mbox{.}}{2019a}]%
        {niu2019knowledge}
\bibfield{author}{\bibinfo{person}{Zhibin Liu}, \bibinfo{person}{Zheng-Yu Niu},
  \bibinfo{person}{Hua Wu}, {and} \bibinfo{person}{Haifeng Wang}.}
  \bibinfo{year}{2019}\natexlab{a}.
\newblock \showarticletitle{Knowledge Aware Conversation Generation with
  Explainable Reasoning over Augmented Graphs}. In
  \bibinfo{booktitle}{\emph{Proceedings of the 2019 Conference on Empirical
  Methods in Natural Language Processing and the 9th International Joint
  Conference on Natural Language Processing (EMNLP-IJCNLP)}}.
  \bibinfo{publisher}{Association for Computational Linguistics},
  \bibinfo{address}{Hong Kong, China}, \bibinfo{pages}{1782--1792}.
\newblock


\bibitem[\protect\citeauthoryear{Mehri and Eskenazi}{Mehri and
  Eskenazi}{2020a}]%
        {mehri2020unsupervised}
\bibfield{author}{\bibinfo{person}{Shikib Mehri} {and} \bibinfo{person}{Maxine
  Eskenazi}.} \bibinfo{year}{2020}\natexlab{a}.
\newblock \showarticletitle{Unsupervised Evaluation of Interactive Dialog with
  {D}ialo{GPT}}. In \bibinfo{booktitle}{\emph{Proceedings of the 21th Annual
  Meeting of the Special Interest Group on Discourse and Dialogue}}.
  \bibinfo{publisher}{Association for Computational Linguistics},
  \bibinfo{address}{1st virtual meeting}, \bibinfo{pages}{225--235}.
\newblock


\bibitem[\protect\citeauthoryear{Mehri and Eskenazi}{Mehri and
  Eskenazi}{2020b}]%
        {mehri2020usr}
\bibfield{author}{\bibinfo{person}{Shikib Mehri} {and} \bibinfo{person}{Maxine
  Eskenazi}.} \bibinfo{year}{2020}\natexlab{b}.
\newblock \showarticletitle{{USR}: An Unsupervised and Reference Free
  Evaluation Metric for Dialog Generation}. In
  \bibinfo{booktitle}{\emph{Proceedings of the 58th Annual Meeting of the
  Association for Computational Linguistics}}. \bibinfo{publisher}{Association
  for Computational Linguistics}, \bibinfo{address}{Online},
  \bibinfo{pages}{681--707}.
\newblock


\bibitem[\protect\citeauthoryear{Meng, Ren, Chen, Monz, Ma, and de~Rijke}{Meng
  et~al\mbox{.}}{2020}]%
        {meng2019refnet}
\bibfield{author}{\bibinfo{person}{Chuan Meng}, \bibinfo{person}{Pengjie Ren},
  \bibinfo{person}{Zhumin Chen}, \bibinfo{person}{Christof Monz},
  \bibinfo{person}{Jun Ma}, {and} \bibinfo{person}{Maarten de Rijke}.}
  \bibinfo{year}{2020}\natexlab{}.
\newblock \showarticletitle{Refnet: A reference-aware network for background
  based conversation}. In \bibinfo{booktitle}{\emph{Proceedings of the AAAI
  Conference on Artificial Intelligence}}, Vol.~\bibinfo{volume}{34}.
  \bibinfo{pages}{8496--8503}.
\newblock


\bibitem[\protect\citeauthoryear{Mockus}{Mockus}{1975}]%
        {mockus1975bayes}
\bibfield{author}{\bibinfo{person}{J Mockus}.} \bibinfo{year}{1975}\natexlab{}.
\newblock \showarticletitle{On the Bayes methods for seeking the extremal
  point}.
\newblock \bibinfo{journal}{\emph{IFAC Proceedings Volumes}}
  \bibinfo{volume}{8}, \bibinfo{number}{1} (\bibinfo{year}{1975}),
  \bibinfo{pages}{428--431}.
\newblock


\bibitem[\protect\citeauthoryear{Pang, Nijkamp, Han, Zhou, Liu, and Tu}{Pang
  et~al\mbox{.}}{2020}]%
        {pang2020towards}
\bibfield{author}{\bibinfo{person}{Bo Pang}, \bibinfo{person}{Erik Nijkamp},
  \bibinfo{person}{Wenjuan Han}, \bibinfo{person}{Linqi Zhou},
  \bibinfo{person}{Yixian Liu}, {and} \bibinfo{person}{Kewei Tu}.}
  \bibinfo{year}{2020}\natexlab{}.
\newblock \showarticletitle{Towards Holistic and Automatic Evaluation of
  Open-Domain Dialogue Generation}. In \bibinfo{booktitle}{\emph{Proceedings of
  the 58th Annual Meeting of the Association for Computational Linguistics}}.
  \bibinfo{publisher}{Association for Computational Linguistics},
  \bibinfo{address}{Online}, \bibinfo{pages}{3619--3629}.
\newblock


\bibitem[\protect\citeauthoryear{Papineni, Roukos, Ward, and Zhu}{Papineni
  et~al\mbox{.}}{2002}]%
        {papineni2002bleu}
\bibfield{author}{\bibinfo{person}{Kishore Papineni}, \bibinfo{person}{Salim
  Roukos}, \bibinfo{person}{Todd Ward}, {and} \bibinfo{person}{Wei-Jing Zhu}.}
  \bibinfo{year}{2002}\natexlab{}.
\newblock \showarticletitle{{B}leu: a Method for Automatic Evaluation of
  Machine Translation}. In \bibinfo{booktitle}{\emph{Proceedings of the 40th
  Annual Meeting of the Association for Computational Linguistics}}.
  \bibinfo{publisher}{Association for Computational Linguistics},
  \bibinfo{address}{Philadelphia, Pennsylvania, USA},
  \bibinfo{pages}{311--318}.
\newblock


\bibitem[\protect\citeauthoryear{Pennington, Socher, and Manning}{Pennington
  et~al\mbox{.}}{2014}]%
        {pennington2014glove}
\bibfield{author}{\bibinfo{person}{Jeffrey Pennington},
  \bibinfo{person}{Richard Socher}, {and} \bibinfo{person}{Christopher
  Manning}.} \bibinfo{year}{2014}\natexlab{}.
\newblock \showarticletitle{{G}lo{V}e: Global Vectors for Word Representation}.
  In \bibinfo{booktitle}{\emph{Proceedings of the 2014 Conference on Empirical
  Methods in Natural Language Processing ({EMNLP})}}.
  \bibinfo{publisher}{Association for Computational Linguistics},
  \bibinfo{address}{Doha, Qatar}, \bibinfo{pages}{1532--1543}.
\newblock


\bibitem[\protect\citeauthoryear{Radford, Wu, Child, Luan, Amodei, and
  Sutskever}{Radford et~al\mbox{.}}{2019}]%
        {radford2019language}
\bibfield{author}{\bibinfo{person}{Alec Radford}, \bibinfo{person}{Jeffrey Wu},
  \bibinfo{person}{Rewon Child}, \bibinfo{person}{David Luan},
  \bibinfo{person}{Dario Amodei}, {and} \bibinfo{person}{Ilya Sutskever}.}
  \bibinfo{year}{2019}\natexlab{}.
\newblock \showarticletitle{Language models are unsupervised multitask
  learners}.
\newblock  (\bibinfo{year}{2019}).
\newblock


\bibitem[\protect\citeauthoryear{Reed, Oraby, and Walker}{Reed
  et~al\mbox{.}}{2018}]%
        {reed2018can}
\bibfield{author}{\bibinfo{person}{Lena Reed}, \bibinfo{person}{Shereen Oraby},
  {and} \bibinfo{person}{Marilyn Walker}.} \bibinfo{year}{2018}\natexlab{}.
\newblock \showarticletitle{Can Neural Generators for Dialogue Learn Sentence
  Planning and Discourse Structuring?}. In
  \bibinfo{booktitle}{\emph{Proceedings of the 11th International Conference on
  Natural Language Generation}}. \bibinfo{pages}{284--295}.
\newblock


\bibitem[\protect\citeauthoryear{Ruder and Plank}{Ruder and Plank}{2017}]%
        {ruder2017learning}
\bibfield{author}{\bibinfo{person}{Sebastian Ruder} {and}
  \bibinfo{person}{Barbara Plank}.} \bibinfo{year}{2017}\natexlab{}.
\newblock \showarticletitle{Learning to select data for transfer learning with
  {B}ayesian Optimization}. In \bibinfo{booktitle}{\emph{Proceedings of the
  2017 Conference on Empirical Methods in Natural Language Processing}}.
  \bibinfo{publisher}{Association for Computational Linguistics},
  \bibinfo{address}{Copenhagen, Denmark}, \bibinfo{pages}{372--382}.
\newblock


\bibitem[\protect\citeauthoryear{See, Roller, Kiela, and Weston}{See
  et~al\mbox{.}}{2019}]%
        {see2019makes}
\bibfield{author}{\bibinfo{person}{Abigail See}, \bibinfo{person}{Stephen
  Roller}, \bibinfo{person}{Douwe Kiela}, {and} \bibinfo{person}{Jason
  Weston}.} \bibinfo{year}{2019}\natexlab{}.
\newblock \showarticletitle{What makes a good conversation? How controllable
  attributes affect human judgments}. In \bibinfo{booktitle}{\emph{Proceedings
  of the 2019 Conference of the North {A}merican Chapter of the Association for
  Computational Linguistics: Human Language Technologies, Volume 1 (Long and
  Short Papers)}}. \bibinfo{publisher}{Association for Computational
  Linguistics}, \bibinfo{address}{Minneapolis, Minnesota},
  \bibinfo{pages}{1702--1723}.
\newblock


\bibitem[\protect\citeauthoryear{Serban, Sordoni, Bengio, Courville, and
  Pineau}{Serban et~al\mbox{.}}{2015}]%
        {serban2015hierarchical}
\bibfield{author}{\bibinfo{person}{Iulian~V Serban},
  \bibinfo{person}{Alessandro Sordoni}, \bibinfo{person}{Yoshua Bengio},
  \bibinfo{person}{Aaron Courville}, {and} \bibinfo{person}{Joelle Pineau}.}
  \bibinfo{year}{2015}\natexlab{}.
\newblock \showarticletitle{Hierarchical neural network generative models for
  movie dialogues}.
\newblock \bibinfo{journal}{\emph{arXiv preprint arXiv:1507.04808}}
  \bibinfo{volume}{7}, \bibinfo{number}{8} (\bibinfo{year}{2015}),
  \bibinfo{pages}{434--441}.
\newblock


\bibitem[\protect\citeauthoryear{Serban, Sordoni, Lowe, Charlin, Pineau,
  Courville, and Bengio}{Serban et~al\mbox{.}}{2017}]%
        {serban2017hierarchical}
\bibfield{author}{\bibinfo{person}{Iulian~Vlad Serban},
  \bibinfo{person}{Alessandro Sordoni}, \bibinfo{person}{Ryan Lowe},
  \bibinfo{person}{Laurent Charlin}, \bibinfo{person}{Joelle Pineau},
  \bibinfo{person}{Aaron~C. Courville}, {and} \bibinfo{person}{Yoshua Bengio}.}
  \bibinfo{year}{2017}\natexlab{}.
\newblock \showarticletitle{A Hierarchical Latent Variable Encoder-Decoder
  Model for Generating Dialogues}. In \bibinfo{booktitle}{\emph{Proceedings of
  the Thirty-First {AAAI} Conference on Artificial Intelligence, February 4-9,
  2017, San Francisco, California, {USA}}},
  \bibfield{editor}{\bibinfo{person}{Satinder~P. Singh} {and}
  \bibinfo{person}{Shaul Markovitch}} (Eds.). \bibinfo{publisher}{{AAAI}
  Press}, \bibinfo{pages}{3295--3301}.
\newblock


\bibitem[\protect\citeauthoryear{Shahriari, Swersky, Wang, Adams, and
  De~Freitas}{Shahriari et~al\mbox{.}}{2015}]%
        {shahriari2015taking}
\bibfield{author}{\bibinfo{person}{Bobak Shahriari}, \bibinfo{person}{Kevin
  Swersky}, \bibinfo{person}{Ziyu Wang}, \bibinfo{person}{Ryan~P Adams}, {and}
  \bibinfo{person}{Nando De~Freitas}.} \bibinfo{year}{2015}\natexlab{}.
\newblock \showarticletitle{Taking the human out of the loop: A review of
  Bayesian optimization}.
\newblock \bibinfo{journal}{\emph{Proc. IEEE}} \bibinfo{volume}{104},
  \bibinfo{number}{1} (\bibinfo{year}{2015}), \bibinfo{pages}{148--175}.
\newblock


\bibitem[\protect\citeauthoryear{Shang, Fu, Peng, Feng, Zhao, and Yan}{Shang
  et~al\mbox{.}}{2018}]%
        {shang2018learning}
\bibfield{author}{\bibinfo{person}{Mingyue Shang}, \bibinfo{person}{Zhenxin
  Fu}, \bibinfo{person}{Nanyun Peng}, \bibinfo{person}{Yansong Feng},
  \bibinfo{person}{Dongyan Zhao}, {and} \bibinfo{person}{Rui Yan}.}
  \bibinfo{year}{2018}\natexlab{}.
\newblock \showarticletitle{Learning to Converse with Noisy Data: Generation
  with Calibration}. In \bibinfo{booktitle}{\emph{Proceedings of the
  Twenty-Seventh International Joint Conference on Artificial Intelligence,
  {IJCAI} 2018, July 13-19, 2018, Stockholm, Sweden}},
  \bibfield{editor}{\bibinfo{person}{J{\'{e}}r{\^{o}}me Lang}} (Ed.).
  \bibinfo{publisher}{ijcai.org}, \bibinfo{pages}{4338--4344}.
\newblock


\bibitem[\protect\citeauthoryear{Shen and Feng}{Shen and Feng}{2020}]%
        {shen2020cdl}
\bibfield{author}{\bibinfo{person}{Lei Shen} {and} \bibinfo{person}{Yang
  Feng}.} \bibinfo{year}{2020}\natexlab{}.
\newblock \showarticletitle{CDL: Curriculum Dual Learning for
  Emotion-Controllable Response Generation}. In
  \bibinfo{booktitle}{\emph{Proceedings of the 58th Annual Meeting of the
  Association for Computational Linguistics}}. \bibinfo{pages}{556--566}.
\newblock


\bibitem[\protect\citeauthoryear{Shen, Feng, and Zhan}{Shen
  et~al\mbox{.}}{2019}]%
        {shen2019modeling}
\bibfield{author}{\bibinfo{person}{Lei Shen}, \bibinfo{person}{Yang Feng},
  {and} \bibinfo{person}{Haolan Zhan}.} \bibinfo{year}{2019}\natexlab{}.
\newblock \showarticletitle{Modeling Semantic Relationship in Multi-turn
  Conversations with Hierarchical Latent Variables}. In
  \bibinfo{booktitle}{\emph{Proceedings of the 57th Annual Meeting of the
  Association for Computational Linguistics}}. \bibinfo{pages}{5497--5502}.
\newblock


\bibitem[\protect\citeauthoryear{Shen, Zhan, Shen, and Feng}{Shen
  et~al\mbox{.}}{2021a}]%
        {shen2021icassp}
\bibfield{author}{\bibinfo{person}{Lei Shen}, \bibinfo{person}{Haolan Zhan},
  \bibinfo{person}{Xin Shen}, {and} \bibinfo{person}{Yang Feng}.}
  \bibinfo{year}{2021}\natexlab{a}.
\newblock \showarticletitle{Learning to select context in a hierarchical and
  global perspective for open-domain dialogue generation}. In
  \bibinfo{booktitle}{\emph{ICASSP 2021-2021 IEEE International Conference on
  Acoustics, Speech and Signal Processing (ICASSP)}}. IEEE,
  \bibinfo{pages}{7438--7442}.
\newblock


\bibitem[\protect\citeauthoryear{Shen, Zhan, Shen, Song, and Zhao}{Shen
  et~al\mbox{.}}{2021b}]%
        {shen2021acmmm}
\bibfield{author}{\bibinfo{person}{Lei Shen}, \bibinfo{person}{Haolan Zhan},
  \bibinfo{person}{Xin Shen}, \bibinfo{person}{Yonghao Song}, {and}
  \bibinfo{person}{Xiaofang Zhao}.} \bibinfo{year}{2021}\natexlab{b}.
\newblock \showarticletitle{Text is NOT Enough: Integrating Visual Impressions
  into Open-domain Dialogue Generation}.
\newblock \bibinfo{journal}{\emph{arXiv preprint arXiv:2109.05778}}
  (\bibinfo{year}{2021}).
\newblock


\bibitem[\protect\citeauthoryear{Sinha, Parthasarathi, Wang, Lowe, Hamilton,
  and Pineau}{Sinha et~al\mbox{.}}{2020}]%
        {sinha2020learning}
\bibfield{author}{\bibinfo{person}{Koustuv Sinha}, \bibinfo{person}{Prasanna
  Parthasarathi}, \bibinfo{person}{Jasmine Wang}, \bibinfo{person}{Ryan Lowe},
  \bibinfo{person}{William~L. Hamilton}, {and} \bibinfo{person}{Joelle
  Pineau}.} \bibinfo{year}{2020}\natexlab{}.
\newblock \showarticletitle{Learning an Unreferenced Metric for Online Dialogue
  Evaluation}. In \bibinfo{booktitle}{\emph{Proceedings of the 58th Annual
  Meeting of the Association for Computational Linguistics}}.
  \bibinfo{publisher}{Association for Computational Linguistics},
  \bibinfo{address}{Online}, \bibinfo{pages}{2430--2441}.
\newblock


\bibitem[\protect\citeauthoryear{Song, Zhang, Hu, and Liu}{Song
  et~al\mbox{.}}{2020}]%
        {song2020generating}
\bibfield{author}{\bibinfo{person}{Haoyu Song}, \bibinfo{person}{Wei-Nan
  Zhang}, \bibinfo{person}{Jingwen Hu}, {and} \bibinfo{person}{Ting Liu}.}
  \bibinfo{year}{2020}\natexlab{}.
\newblock \showarticletitle{Generating persona consistent dialogues by
  exploiting natural language inference}. In
  \bibinfo{booktitle}{\emph{Proceedings of the AAAI Conference on Artificial
  Intelligence}}, Vol.~\bibinfo{volume}{34}. \bibinfo{pages}{8878--8885}.
\newblock


\bibitem[\protect\citeauthoryear{Tao, Mou, Zhao, and Yan}{Tao
  et~al\mbox{.}}{2018}]%
        {tao2018ruber}
\bibfield{author}{\bibinfo{person}{Chongyang Tao}, \bibinfo{person}{Lili Mou},
  \bibinfo{person}{Dongyan Zhao}, {and} \bibinfo{person}{Rui Yan}.}
  \bibinfo{year}{2018}\natexlab{}.
\newblock \showarticletitle{{RUBER:} An Unsupervised Method for Automatic
  Evaluation of Open-Domain Dialog Systems}. In
  \bibinfo{booktitle}{\emph{Proceedings of the Thirty-Second {AAAI} Conference
  on Artificial Intelligence, (AAAI-18), the 30th innovative Applications of
  Artificial Intelligence (IAAI-18), and the 8th {AAAI} Symposium on
  Educational Advances in Artificial Intelligence (EAAI-18), New Orleans,
  Louisiana, USA, February 2-7, 2018}},
  \bibfield{editor}{\bibinfo{person}{Sheila~A. McIlraith} {and}
  \bibinfo{person}{Kilian~Q. Weinberger}} (Eds.). \bibinfo{publisher}{{AAAI}
  Press}, \bibinfo{pages}{722--729}.
\newblock


\bibitem[\protect\citeauthoryear{Tsvetkov, Faruqui, Ling, MacWhinney, and
  Dyer}{Tsvetkov et~al\mbox{.}}{2016}]%
        {tsvetkov2016learning}
\bibfield{author}{\bibinfo{person}{Yulia Tsvetkov}, \bibinfo{person}{Manaal
  Faruqui}, \bibinfo{person}{Wang Ling}, \bibinfo{person}{Brian MacWhinney},
  {and} \bibinfo{person}{Chris Dyer}.} \bibinfo{year}{2016}\natexlab{}.
\newblock \showarticletitle{Learning the Curriculum with {B}ayesian
  Optimization for Task-Specific Word Representation Learning}. In
  \bibinfo{booktitle}{\emph{Proceedings of the 54th Annual Meeting of the
  Association for Computational Linguistics (Volume 1: Long Papers)}}.
  \bibinfo{publisher}{Association for Computational Linguistics},
  \bibinfo{address}{Berlin, Germany}, \bibinfo{pages}{130--139}.
\newblock


\bibitem[\protect\citeauthoryear{Vaswani, Shazeer, Parmar, Uszkoreit, Jones,
  Gomez, Kaiser, and Polosukhin}{Vaswani et~al\mbox{.}}{2017}]%
        {vaswani2017attention}
\bibfield{author}{\bibinfo{person}{Ashish Vaswani}, \bibinfo{person}{Noam
  Shazeer}, \bibinfo{person}{Niki Parmar}, \bibinfo{person}{Jakob Uszkoreit},
  \bibinfo{person}{Llion Jones}, \bibinfo{person}{Aidan~N. Gomez},
  \bibinfo{person}{Lukasz Kaiser}, {and} \bibinfo{person}{Illia Polosukhin}.}
  \bibinfo{year}{2017}\natexlab{}.
\newblock \showarticletitle{Attention is All you Need}. In
  \bibinfo{booktitle}{\emph{Advances in Neural Information Processing Systems
  30: Annual Conference on Neural Information Processing Systems 2017, December
  4-9, 2017, Long Beach, CA, {USA}}},
  \bibfield{editor}{\bibinfo{person}{Isabelle Guyon}, \bibinfo{person}{Ulrike
  von Luxburg}, \bibinfo{person}{Samy Bengio}, \bibinfo{person}{Hanna~M.
  Wallach}, \bibinfo{person}{Rob Fergus}, \bibinfo{person}{S.~V.~N.
  Vishwanathan}, {and} \bibinfo{person}{Roman Garnett}} (Eds.).
  \bibinfo{pages}{5998--6008}.
\newblock


\bibitem[\protect\citeauthoryear{Walker, Litman, Kamm, and Abella}{Walker
  et~al\mbox{.}}{1997}]%
        {walker1997paradise}
\bibfield{author}{\bibinfo{person}{Marilyn Walker}, \bibinfo{person}{Diane
  Litman}, \bibinfo{person}{Candace~A Kamm}, {and} \bibinfo{person}{Alicia
  Abella}.} \bibinfo{year}{1997}\natexlab{}.
\newblock \showarticletitle{PARADISE: A Framework for Evaluating Spoken
  Dialogue Agents}. In \bibinfo{booktitle}{\emph{35th Annual Meeting of the
  Association for Computational Linguistics and 8th Conference of the European
  Chapter of the Association for Computational Linguistics}}.
  \bibinfo{pages}{271--280}.
\newblock


\bibitem[\protect\citeauthoryear{Welleck, Weston, Szlam, and Cho}{Welleck
  et~al\mbox{.}}{2019}]%
        {welleck2019dialogue}
\bibfield{author}{\bibinfo{person}{Sean Welleck}, \bibinfo{person}{Jason
  Weston}, \bibinfo{person}{Arthur Szlam}, {and} \bibinfo{person}{Kyunghyun
  Cho}.} \bibinfo{year}{2019}\natexlab{}.
\newblock \showarticletitle{Dialogue Natural Language Inference}. In
  \bibinfo{booktitle}{\emph{Proceedings of the 57th Annual Meeting of the
  Association for Computational Linguistics}}. \bibinfo{publisher}{Association
  for Computational Linguistics}, \bibinfo{address}{Florence, Italy},
  \bibinfo{pages}{3731--3741}.
\newblock


\bibitem[\protect\citeauthoryear{Xu, Du{\v{s}}ek, Konstas, and Rieser}{Xu
  et~al\mbox{.}}{2018}]%
        {xu2018better}
\bibfield{author}{\bibinfo{person}{Xinnuo Xu}, \bibinfo{person}{Ond{\v{r}}ej
  Du{\v{s}}ek}, \bibinfo{person}{Ioannis Konstas}, {and}
  \bibinfo{person}{Verena Rieser}.} \bibinfo{year}{2018}\natexlab{}.
\newblock \showarticletitle{Better Conversations by Modeling, Filtering, and
  Optimizing for Coherence and Diversity}. In
  \bibinfo{booktitle}{\emph{Proceedings of the 2018 Conference on Empirical
  Methods in Natural Language Processing}}. \bibinfo{publisher}{Association for
  Computational Linguistics}, \bibinfo{address}{Brussels, Belgium},
  \bibinfo{pages}{3981--3991}.
\newblock


\bibitem[\protect\citeauthoryear{Zhan, Zhang, Chen, Ding, Bao, and Lan}{Zhan
  et~al\mbox{.}}{2021a}]%
        {zhan2021naacl}
\bibfield{author}{\bibinfo{person}{Haolan Zhan}, \bibinfo{person}{Hainan
  Zhang}, \bibinfo{person}{Hongshen Chen}, \bibinfo{person}{Zhuoye Ding},
  \bibinfo{person}{Yongjun Bao}, {and} \bibinfo{person}{Yanyan Lan}.}
  \bibinfo{year}{2021}\natexlab{a}.
\newblock \showarticletitle{Augmenting knowledge-grounded conversations with
  sequential knowledge transition}. In \bibinfo{booktitle}{\emph{Proceedings of
  the 2021 Conference of the North American Chapter of the Association for
  Computational Linguistics: Human Language Technologies}}.
  \bibinfo{pages}{5621--5630}.
\newblock


\bibitem[\protect\citeauthoryear{Zhan, Zhang, Chen, Ding, Bao, and Lan}{Zhan
  et~al\mbox{.}}{2021b}]%
        {zhan2021augmenting}
\bibfield{author}{\bibinfo{person}{Haolan Zhan}, \bibinfo{person}{Hainan
  Zhang}, \bibinfo{person}{Hongshen Chen}, \bibinfo{person}{Zhuoye Ding},
  \bibinfo{person}{Yongjun Bao}, {and} \bibinfo{person}{Yanyan Lan}.}
  \bibinfo{year}{2021}\natexlab{b}.
\newblock \showarticletitle{Augmenting knowledge-grounded conversations with
  sequential knowledge transition}. In \bibinfo{booktitle}{\emph{Proceedings of
  the 2021 Conference of the North American Chapter of the Association for
  Computational Linguistics: Human Language Technologies}}.
  \bibinfo{pages}{5621--5630}.
\newblock


\bibitem[\protect\citeauthoryear{Zhang, Galley, Gao, Gan, Li, Brockett, and
  Dolan}{Zhang et~al\mbox{.}}{2018}]%
        {zhang2018generating}
\bibfield{author}{\bibinfo{person}{Yizhe Zhang}, \bibinfo{person}{Michel
  Galley}, \bibinfo{person}{Jianfeng Gao}, \bibinfo{person}{Zhe Gan},
  \bibinfo{person}{Xiujun Li}, \bibinfo{person}{Chris Brockett}, {and}
  \bibinfo{person}{Bill Dolan}.} \bibinfo{year}{2018}\natexlab{}.
\newblock \showarticletitle{Generating Informative and Diverse Conversational
  Responses via Adversarial Information Maximization}. In
  \bibinfo{booktitle}{\emph{Advances in Neural Information Processing Systems
  31: Annual Conference on Neural Information Processing Systems 2018, NeurIPS
  2018, December 3-8, 2018, Montr{\'{e}}al, Canada}},
  \bibfield{editor}{\bibinfo{person}{Samy Bengio}, \bibinfo{person}{Hanna~M.
  Wallach}, \bibinfo{person}{Hugo Larochelle}, \bibinfo{person}{Kristen
  Grauman}, \bibinfo{person}{Nicol{\`{o}} Cesa{-}Bianchi}, {and}
  \bibinfo{person}{Roman Garnett}} (Eds.). \bibinfo{pages}{1815--1825}.
\newblock


\bibitem[\protect\citeauthoryear{Zhao, Lala, and Kawahara}{Zhao
  et~al\mbox{.}}{2020}]%
        {zhao2020designing}
\bibfield{author}{\bibinfo{person}{Tianyu Zhao}, \bibinfo{person}{Divesh Lala},
  {and} \bibinfo{person}{Tatsuya Kawahara}.} \bibinfo{year}{2020}\natexlab{}.
\newblock \showarticletitle{Designing Precise and Robust Dialogue Response
  Evaluators}. In \bibinfo{booktitle}{\emph{Proceedings of the 58th Annual
  Meeting of the Association for Computational Linguistics}}.
  \bibinfo{publisher}{Association for Computational Linguistics},
  \bibinfo{address}{Online}, \bibinfo{pages}{26--33}.
\newblock


\bibitem[\protect\citeauthoryear{Zhao, Zhao, and Eskenazi}{Zhao
  et~al\mbox{.}}{2017}]%
        {zhao2017learning}
\bibfield{author}{\bibinfo{person}{Tiancheng Zhao}, \bibinfo{person}{Ran Zhao},
  {and} \bibinfo{person}{Maxine Eskenazi}.} \bibinfo{year}{2017}\natexlab{}.
\newblock \showarticletitle{Learning Discourse-level Diversity for Neural
  Dialog Models using Conditional Variational Autoencoders}. In
  \bibinfo{booktitle}{\emph{Proceedings of the 55th Annual Meeting of the
  Association for Computational Linguistics (Volume 1: Long Papers)}}.
  \bibinfo{publisher}{Association for Computational Linguistics},
  \bibinfo{address}{Vancouver, Canada}, \bibinfo{pages}{654--664}.
\newblock


\bibitem[\protect\citeauthoryear{Zhou, Huang, Zhang, Zhu, and Liu}{Zhou
  et~al\mbox{.}}{2018a}]%
        {zhou2018emotional}
\bibfield{author}{\bibinfo{person}{Hao Zhou}, \bibinfo{person}{Minlie Huang},
  \bibinfo{person}{Tianyang Zhang}, \bibinfo{person}{Xiaoyan Zhu}, {and}
  \bibinfo{person}{Bing Liu}.} \bibinfo{year}{2018}\natexlab{a}.
\newblock \showarticletitle{Emotional Chatting Machine: Emotional Conversation
  Generation with Internal and External Memory}. In
  \bibinfo{booktitle}{\emph{Proceedings of the Thirty-Second {AAAI} Conference
  on Artificial Intelligence, (AAAI-18), the 30th innovative Applications of
  Artificial Intelligence (IAAI-18), and the 8th {AAAI} Symposium on
  Educational Advances in Artificial Intelligence (EAAI-18), New Orleans,
  Louisiana, USA, February 2-7, 2018}},
  \bibfield{editor}{\bibinfo{person}{Sheila~A. McIlraith} {and}
  \bibinfo{person}{Kilian~Q. Weinberger}} (Eds.). \bibinfo{publisher}{{AAAI}
  Press}, \bibinfo{pages}{730--739}.
\newblock


\bibitem[\protect\citeauthoryear{Zhou, Young, Huang, Zhao, Xu, and Zhu}{Zhou
  et~al\mbox{.}}{2018b}]%
        {zhou2018commonsense}
\bibfield{author}{\bibinfo{person}{Hao Zhou}, \bibinfo{person}{Tom Young},
  \bibinfo{person}{Minlie Huang}, \bibinfo{person}{Haizhou Zhao},
  \bibinfo{person}{Jingfang Xu}, {and} \bibinfo{person}{Xiaoyan Zhu}.}
  \bibinfo{year}{2018}\natexlab{b}.
\newblock \showarticletitle{Commonsense Knowledge Aware Conversation Generation
  with Graph Attention}. In \bibinfo{booktitle}{\emph{Proceedings of the
  Twenty-Seventh International Joint Conference on Artificial Intelligence,
  {IJCAI} 2018, July 13-19, 2018, Stockholm, Sweden}},
  \bibfield{editor}{\bibinfo{person}{J{\'{e}}r{\^{o}}me Lang}} (Ed.).
  \bibinfo{publisher}{ijcai.org}, \bibinfo{pages}{4623--4629}.
\newblock


\end{thebibliography}

\end{document}